\documentclass[11pt,a4paper]{article}
\usepackage{amssymb}
\usepackage{graphicx}
\usepackage{caption}
\usepackage{subcaption}
\usepackage[ruled,vlined]{algorithm2e}
\usepackage{url}
\usepackage{enumerate}
\usepackage{amsmath}
\usepackage{color}
\usepackage{todonotes}
\usepackage{authblk}
\usepackage{lscape}

\newtheorem{definition}{Definition}

\newtheorem{example}{Example}

\begin{document}

\title{Probabilistic Relational Model Benchmark Generation}

\author{
\small Mouna Ben Ishak, Rajani Chulyadyo, and Philippe Leray \\
\small LARODEC Laboratory, ISG, Universit\'e de Tunis, Tunisia\\
\small DUKe research group, LINA Laboratory UMR 6241, University of Nantes, France\\
\small DataForPeople, Nantes, France}
\date{}

\maketitle

\begin{abstract}
The validation of any database mining methodology goes through an evaluation process where benchmarks availability is essential. In this paper, we aim to randomly generate relational database benchmarks that allow to check probabilistic dependencies among the attributes. We are particularly interested in Probabilistic Relational Models (PRMs), which extend Bayesian Networks (BNs) to a relational data mining context and enable effective and robust reasoning over relational data. Even though a panoply of works have focused, separately, on the generation of random Bayesian networks and relational databases, no work has been identified for PRMs on that track. This paper provides an algorithmic approach for generating random PRMs from scratch to fill this gap. The proposed method allows to generate PRMs as well as synthetic relational data from a randomly generated relational schema and a random set of probabilistic dependencies. This can be of interest not only for machine learning researchers to evaluate their proposals in a common framework, but also for databases designers to evaluate the effectiveness of the components of a database management system.\\

\end{abstract}

\section{Introduction}
Data mining is the central step in knowledge discovery in databases. It relies on several research areas including statistics and machine learning. Usually, machine learning techniques are developed around flat data representation (i.e., matrix form) and are known as propositional learning approaches. However, due to the development of communication and storage technologies, data management practices have taken further aspects. Data can present a very large number of dimensions, with several different types of entities. With the growing interest in extracting patterns from such data representation, relational data mining approaches have emerged with the interest of finding patterns in a given relational database~\cite{datamining01} and Statistical Relational Learning (SRL) has emerged as an area of machine learning that enables effective and robust reasoning about relational data structures~\cite{SL07}. In this paper, we are particularly interested in Probabilistic Relational Models (PRMs)\footnote{Neville and Jensen~\cite{Jensen07} use the term `Relational Bayesian Network' to refer to Bayesian networks that have been extended to model relational databases~\cite{Koller98,Pfeffer00}
and use the term `PRM' in its more general sense to distinguish the family of probabilistic graphical models that are interested in extracting statistical patterns from relational models. In this paper, we preserve the term PRM as used by~\cite{Koller98,Pfeffer00}.}~\cite{Koller98,Pfeffer00}, which represent a relational extension of Bayesian networks~\cite{Pearl88}, where the probability model specification concerns classes of objects rather than simple attributes. PRMs present a probabilistic graphical formalism that enables flexible modeling of complex relational interactions.
	
PRMs have proved their applicability in several areas (e.g., risk analysis, web page classification, recommender systems)~\cite{rajani14,Fersini09,Teodor10} as they allow to minimize data preprocessing and the loss of significant information~\cite{Raedt98}. The use of PRMs implies their construction either by experts or by applying learning algorithms in order to learn the model from some existing observational relational data. PRMs learning involves finding a graphical structure as well as a set of conditional probability distributions that best fit to the relational training data. The evaluation of the learning approaches is usually done using randomly generated data coming from either real known networks or randomly generated ones. However, neither the first nor the second are available in the literature. Moreover, there is a growing interest from the database community to produce database benchmarks to support and illustrate decision support systems (DSSs). For real-world business tasks, uncertainty is an unmissable aspect. So, benchmarks designed to support DSSs should consider this task. 

In this paper, we propose an algorithmic approach that allows to generate random PRMs from scratch, and then populate a database instance. The originality of this process is that it allows to generate synthetic relational data from a randomly generated relational schema and a random set of probabilistic dependencies. Since PRMs bring together two neighboring subfields of computer science, namely machine learning and database management, our process can be useful for both domains. It is imperative for statistical relational learning researchers to evaluate the effectiveness of their learning approaches. On the other hand, it can be of interest for database designers to evaluate the effectiveness of a database management system (DBMS) components. It allows to generate various relational schemas, from simple to complex ones, and to populate database tables with huge number of tuples derived from underlying probability distributions defined by the generated PRMs. This paper presents an extended version of a preliminary work published in~\cite{mouna14,BenIshakIDA}.

\section{Background}
This section first provides a brief recall on Bayesian networks and relational model, and then introduces PRMs.

\subsection{Bayesian networks}
\label{BN}
Bayesian networks (BNs)~\cite{Pearl88} are directed acyclic graphs allowing to efficiently encode and manipulate probability distributions over high-dimensional spaces. Formally, they are defined as follows: 

\begin{definition}
A Bayesian network $B = \left\langle \mathcal G,\Theta\right\rangle $ is defined by:

\begin{enumerate}[{1)}]
	\item A graphical component (structure): a directed acyclic graph (DAG) $\mathcal G = \left(V,E\right)$, where $V$ is the set of vertices representing $n$ discrete random variables $\mathcal A = \left\{A_{1},\ldots,A_{n}\right\}$, and $E$ is the set of directed edges corresponding to conditional dependence relationships among these variables.
 \item A numerical component (parameters): $\Theta = \left\{\Theta_{1},\ldots,\Theta_{n}\right\}$ where each $\Theta_{i} = P\left(A_{i}|Pa\left(A_{i}\right)\right)$ denotes the conditional probability distribution (CPD) of each node $A_{i}$ given its parents in $\mathcal G$ denoted by $Pa\left(A_{i}\right)$.
\end{enumerate}
\end{definition}

Several approaches have been proposed to learn BNs from data~\cite{daly11}. The evaluation of these learning algorithms requires either the use of known networks or the use of a random generation process. The former allows to sample data and perform learning using this data in order to recover the initial gold standard net. The latter allows to generate synthetic BNs and data in order to provide a large number of possible models and to carry out experimentation while varying models from simple to complex ones.\\

\textbf{Random Bayesian networks generation} comes to provide a graph structure and parameters. Statnikov et al.~\cite{Tsamardinos03} proposed an algorithmic approach to generate arbitrarily large BNs by tiling smaller real-world known networks. The complexity of the final model is controlled by the number of tiling and a connectivity parameter which determines the maximum number of connections between one node and the next tile. Some works have been devoted to the generation of synthetic networks but without any guarantee that every allowed graph is produced with the same uniform probability~\cite{Ide04generatingrandom}. In~\cite{Ide02randomgeneration}, the authors have proposed an approach, called PMMixed algorithm, that allows the generation of uniformly distributed Bayesian networks using Markov chains. Using this algorithm, constraints on generated nets can be added with relative ease such as constraints on nodes degree, maximum number of dependencies in the graph, etc. Once the DAG structure is generated, it is easy to construct a complete Bayesian network by randomly generating associated probability distributions by sampling either Uniform or Dirichlet distributions. Having the final BN, standard sampling method, such as forward sampling~\cite{Henrion86}, can be used to generate observational data.

\subsection{Relational model}
\label{RBD}
The manner how the data is organized in a database depends on the chosen database model. The relational model is the most commonly used one and it represents the basis for the most large scale knowledge representation systems~\cite{datamining01}. Formally, the relational representation can be defined as follows: 

\begin{definition}
The relational representation consists of
\begin{itemize}
	\item A set of relations (or tables or classes) $\mathcal X = \{X_{1},\ldots,X_{n}\}$. Each relation $X_{i}$ has two parts: 
\begin{itemize}
	\item The heading (relation schema): a fixed set of attributes $\mathcal A(X)=\{A_{1},\ldots,A_{k}\}$. Each attribute $A_{i}$ is characterized by a name and a domain denoted $D_{i}$. 
	\item The body: a set of tuples (or records). Each tuple associates for each attribute $A_{i}$ in the heading a value from its domain $D_{i}$.
\end{itemize}
 \item Each relation has a key (i.e., a unique identifier, a subset of the heading of a relation $X_{i}$.) and, possibly, a set of foreign key attributes (or \textbf{reference slots $\rho$}). A foreign key attribute is a field that points to a key field in another relation, called the referenced relation. The associated constraint is a \textbf{referential constraint}. A chain of such constraints constitutes a \textbf{referential path}. If a referential path from some relation to itself is found then it is called a \textbf{referential cycle}\footnote{Database designs involving referential cycles are usually contraindicated~\cite{Date08}.}. Relation headings and constraints are described by a \textbf{relational schema $\mathcal R$}. 
\end{itemize}
\end{definition}

Usually the interaction with a relational database is ensured by specifying queries using structured query language (SQL), which on their part use some specific operators to extract significant meaning such as \textbf{aggregators}. An aggregation function $\gamma$ takes a multi-set of values of some ground type, and returns a summary of it. Some requests need to cross long reference paths, with some possible back and forth. They use composed slots to define functions from some objects to other ones to which they are indirectly related. We call this composition of slots a \textbf{slot chain $K$}. We call a slot chain single-valued when all the crossed reference slots end with a cardinality equal to $1$. A slot chain is multi-valued if it contains at least one reference slot ending with cardinality equal to $many$. Multi-valued slot chains imply the use of aggregators.

Generally, database benchmarks are used to measure the performance of a database system. A database benchmark includes several subtasks (e.g., generating the transaction workload, defining transaction logic, generating the database)~\cite{Gray92}.\\

\textbf{Random database generation} consists on creating the database schema, determining data distribution, generating it and loading all these components to the database system under test. Several propositions have been developed in this context. The main issue was how to provide a large number of records using some known distributions in order to be able to evaluate the system results~\cite{Bitton83,Bruno05}. In some research, known benchmarks~\footnote{http://www.tpc.org} are used and the ultimate goal is only to generate a large dataset~\cite{Gray94}. Nowadays, several software tools are available (e.g., \textit{DbSchema}\footnote{http://www.dbschema.com/}, \textit{DataFiller}\footnote{https://www.cri.ensmp.fr/people/coelho/datafiller.html}) to populate database instances knowing the relational schema structure. Records are then generated on the basis of this input by considering that the attributes are probabilistically independent which is not relevant when these benchmarks are used to evaluate decision support systems. The Transaction Processing Performance Council (TPC)\footnote{http://www.tpc.org} organization provides the TPC-DS\footnote{http://www.tpc.org/tpcds} benchmark which has been designed to be suitable with real-world business tasks which are characterized by the analysis of huge amount of data. The TPC-DS schema models sales and the sales returns process for an organization. TPC-DS provides tools to generate either data sets or query sets for the benchmark. Nevertheless, uncertainty management stays a prominent challenge to provide better rational decision making.

\subsection{Probabilistic relational models}
\label{PRM}
Probabilistic relational models~\cite{SL07Chap5,Koller98,Pfeffer00} are an extension of BNs in the relational context. They bring together the strengths of probabilistic graphical models and the relational presentation. Formally, they are defined as follows~\cite{SL07Chap5}: 

\begin{definition}
\label{def3}
A Probabilistic Relational Model $\Pi$ for a relational schema $\mathcal R$ is defined by:

\begin{enumerate}[{1)}]
	\item A qualitative dependency structure $\mathcal S:$ for each class (relation) $X \in \mathcal X$ and each descriptive attribute $A \in \mathcal A(X)$, there is a set of parents $Pa(X.A)=\{U_{1},\ldots,U_{l}\}$ that describes probabilistic dependencies. Each $U_{i}$ has the form $X.B$ if it is a simple attribute in the same relation or $\gamma(X.K.B)$, where $K$ is a slot chain and $\gamma$ is an aggregation function.
	\item A quantitative component, a set of conditional probability distributions (CPDs), representing $P(X.A | Pa(X.A))$.
\end{enumerate}
\end{definition}

The PRM $\Pi$ is a meta-model used to describe the overall behavior of a system. To perform probabilistic inference, this model has to be instantiated. A PRM instance contains, for each class of $\Pi$, the set of objects involved in the system and the relations that hold between them (i.e., tuples from the database instance which are interlinked). This structure is known as a relational skeleton $\sigma_{r}$~\cite{SL07Chap5}.

\begin{definition}
\label{defskeleton}
A relational skeleton $\sigma_{r}$ of a relational schema is a partial specification of an instance of the schema. It specifies the set of objects $\sigma_{r}(Xi)$ for each class and the relations that hold between the objects. However, it leaves the values of the
attributes unspecified.
\end{definition}

Given a relational skeleton, the PRM $\Pi$ defines a distribution over the possible worlds consistent with $\sigma_{r}$ through a ground Bayesian network~\cite{SL07Chap5}. 

\begin{definition}
\label{def4}
A Ground Bayesian Network (GBN) is defined given a PRM $\Pi$ together with a relational skeleton $\sigma_{r}$. A GBN consists of:
\begin{enumerate}[{1)}]
	\item A qualitative component:
\begin{itemize}
	\item A node for every attribute of every object $x\in \sigma_{r}(X), x.A$.
	\item Each $x.A$ depends probabilistically on a set of parents $Pa(x.A)={u_{1},\ldots u_{l}}$ of the form $x.B$ or $x.K.B$, where each $u_{i}$ is an instance of the $U_{i}$ defined in the PRM. If $K$ is not single-valued, then the parent is an aggregate computed from the set of random variables $\{y | y \in x.K\}, \gamma(x.K.B)$.
\end{itemize}
	\item A quantitative component, the CPD for $x.A$ is $P(X.A|Pa(X.A))$ .
\end{enumerate}
\end{definition}

\begin{example}

\begin{figure}[!t]
\centering
	\includegraphics[width=0.65\textwidth]{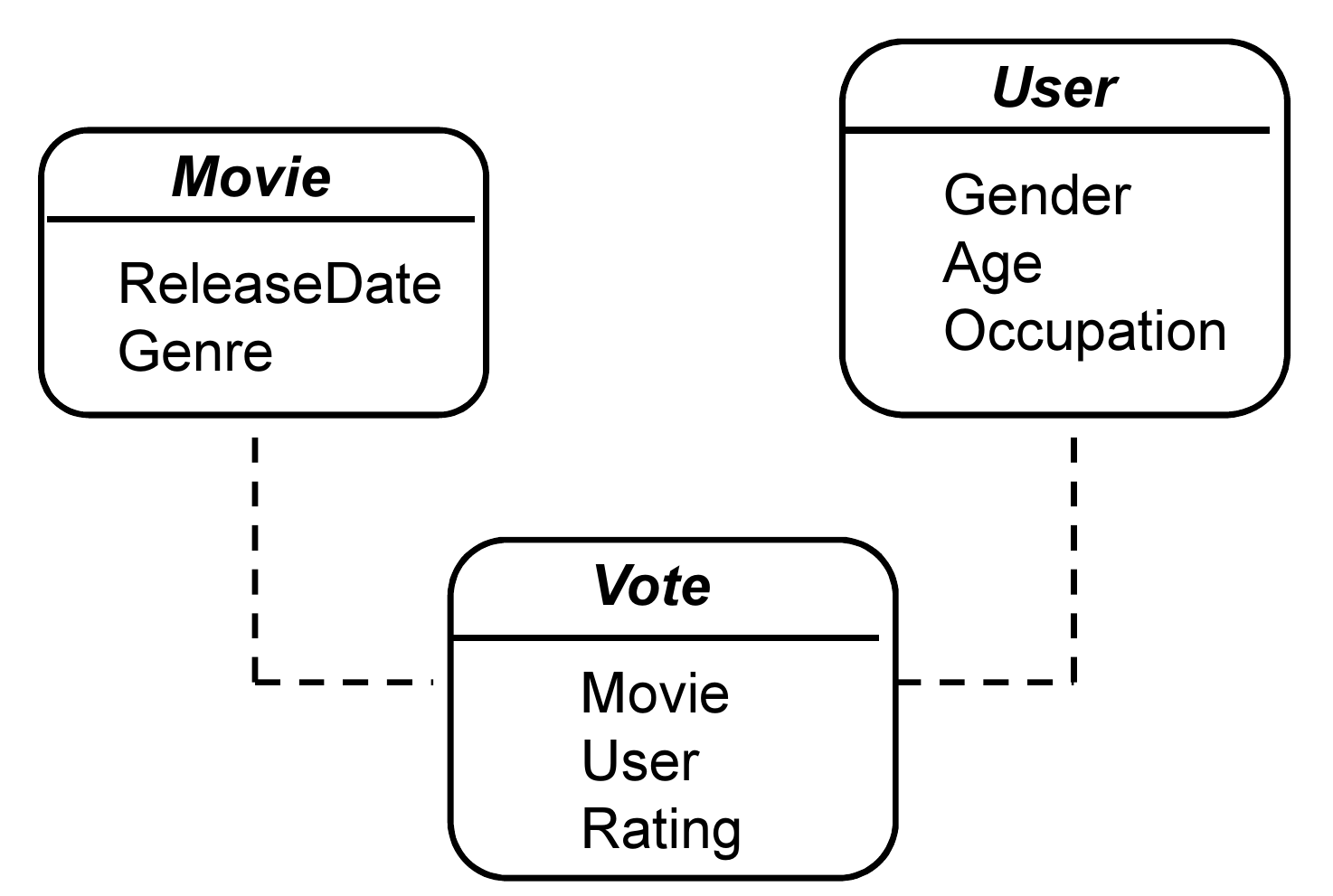}
	\caption{An example of relational schema}
	\label{subfig1}
\end{figure}

\begin{figure}[!t]
\centering
	\includegraphics[width=0.95\textwidth]{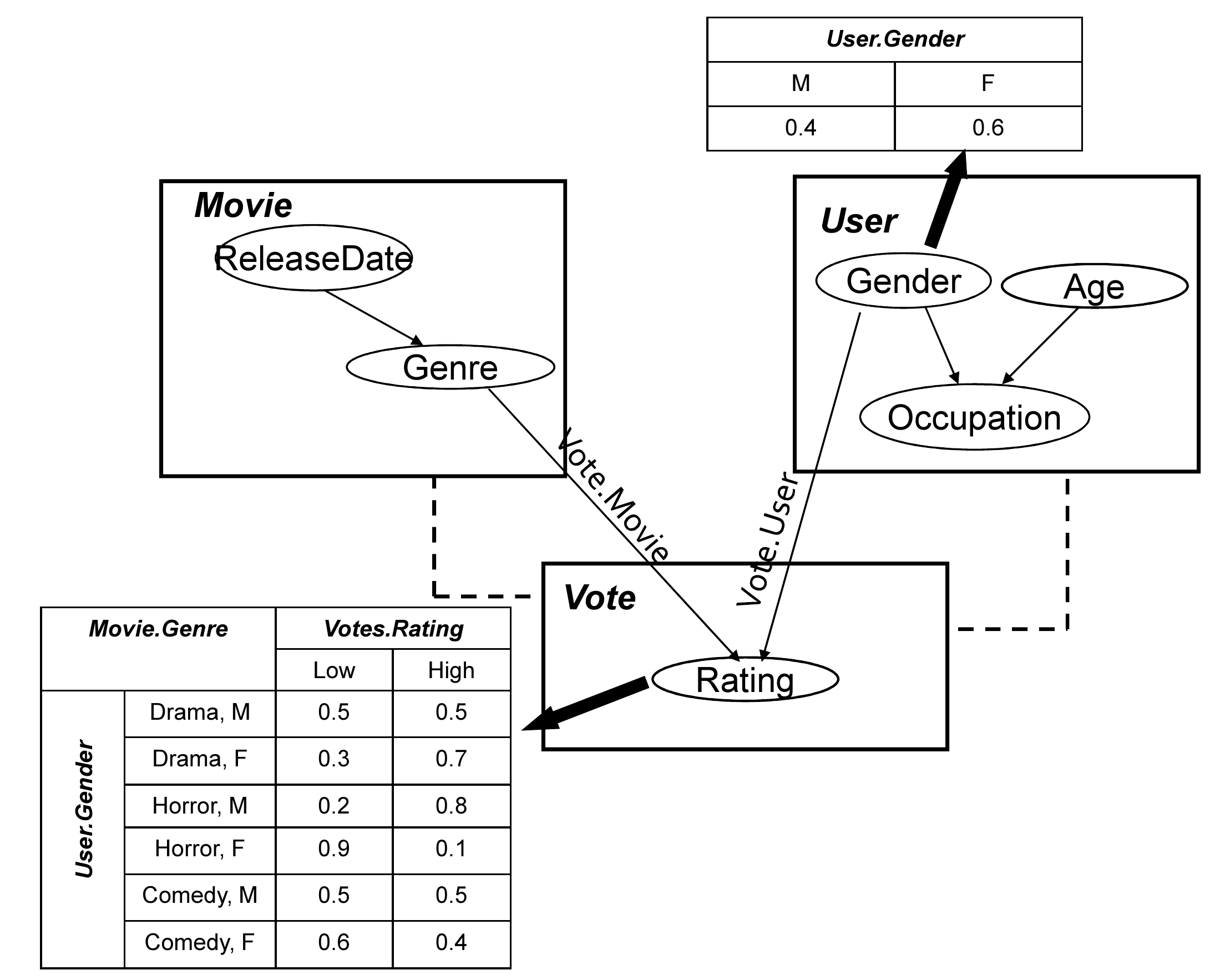}
	\caption{An example of probabilistic relational model}
	\label{subfig2}
\end{figure}

An example of a relational schema is depicted in Figure~\ref{subfig1}, with three classes $\mathcal{X} = \{Movie, Vote, User\}$. The relation $Vote$ has a descriptive attribute $Vote.Rating$ and two reference slots $Vote.User$ and $Vote.Movie$. $Vote.User$ relates the objects of class $Vote$ with the objects of class $User$. Dotted links presents reference slots.
An example of a slot chain would be $Vote.User.User^{-1}.Movie$ which could be interpreted as all the votes of movies cast by a particular user.\\
$Vote.Movie.genre \rightarrow Vote.rating$ is an example of a probabilistic dependency derived from a slot chain of length $1$ where $Vote.Movie.genre$ is a parent of $Vote.rating$ as shown in Figure~\ref{subfig2}. Also, varying the slot chain length may give rise to other dependencies. For instance, using a slot chain of length $3$, we can have a probabilistic dependency from $\gamma(Vote.User.User^{-1}.Movie.genre)$ to $Vote.rating$. In this case, $Vote.rating$ depends probabilistically on an aggregate value of all the genres of movies voted by a particular user. 

\begin{figure}[!t]
\centering
	\includegraphics[width=0.7\textwidth]{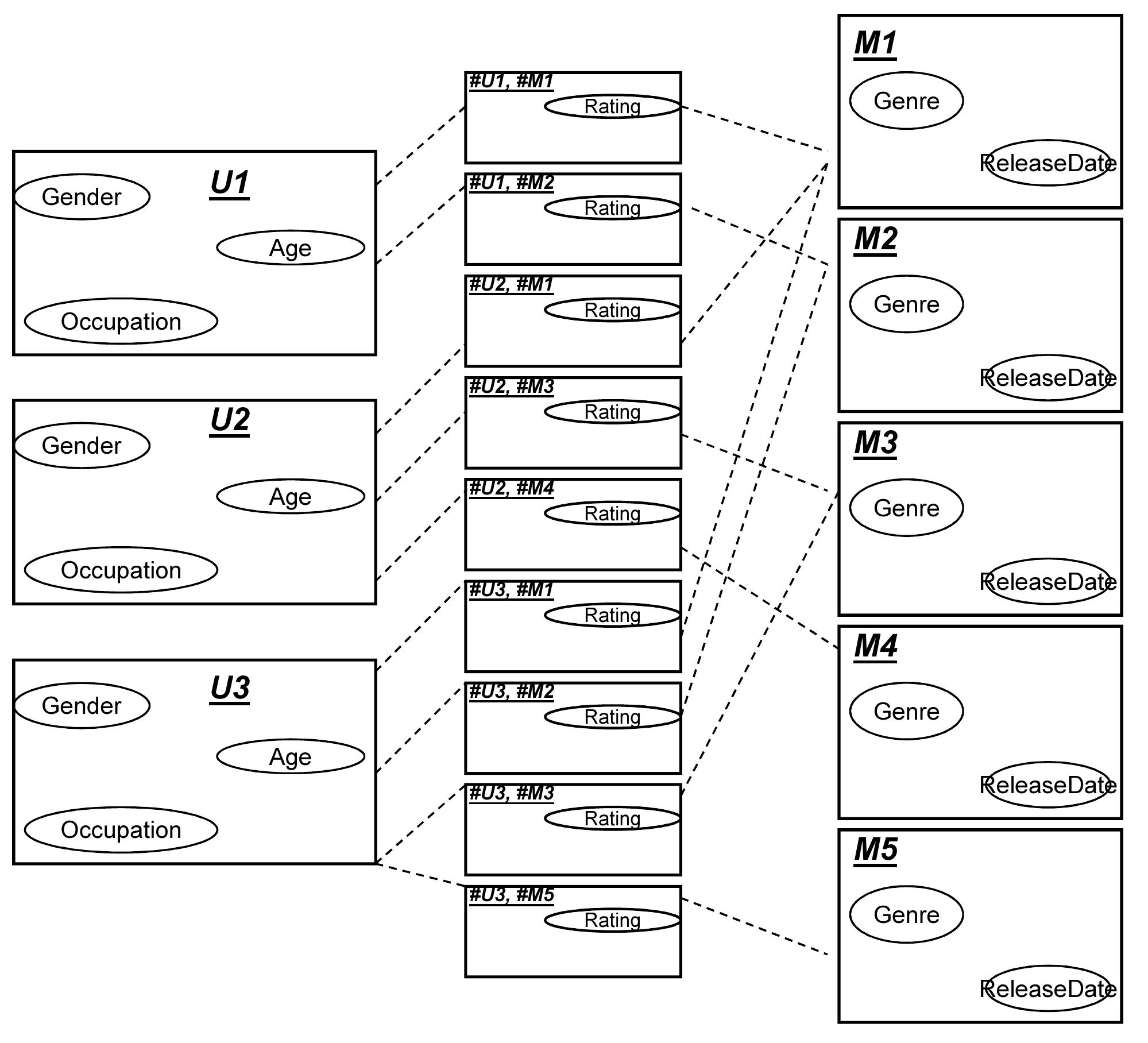}
	\caption{An example of a relational skeleton}
	\label{subfig3}
\end{figure}

Figure~\ref{subfig3} is an example of a relational skeleton of the relational schema of Figure~\ref{subfig1}. This relational skeleton contains $3$ users, $5$ movies and $9$ votes. Also it specifies the relations between these objects, e.g., the user $U1$ voted for two movies $m1$ and $m2$.

\begin{figure}[!t]
\centering
	\includegraphics[width=0.7\textwidth]{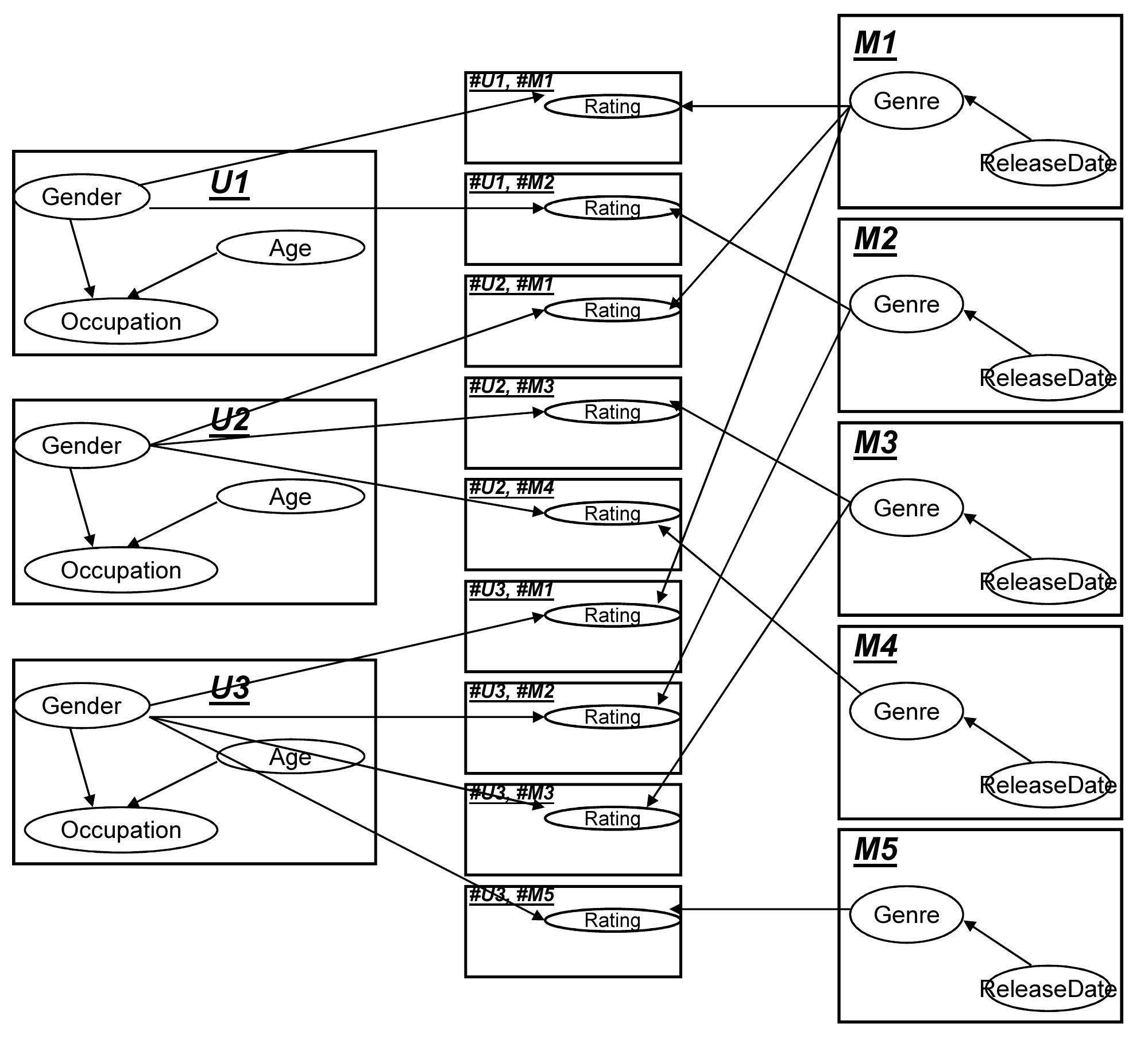}
	\caption{An example of a ground Bayesian network}
	\label{subfig4}
\end{figure}

Figure~\ref{subfig4} presents the ground Bayesian network constructed from the relational skeleton of Figure~\ref{subfig3} and the PRM of Figure~\ref{subfig2}. It resumes the same dependencies as well as $CPDs$ of the PRM at the level of objects. Here, we have not reproduced the $CPDs$ to not overload the figure.

\end{example}

\textbf{PRM structure learning} has not been well studied in the literature. Only few works have been proposed to learn PRMs~\cite{Friedman99} or almost similar models~\cite{Maier132,Maier10} from relational data.\\

Friedman et al.~\cite{Friedman99} proposed Relational Greedy Hill-Climbing Search (RGS) algorithm. They applied a greedy search procedure to explore the space of PRM structures while allowing increasingly large slot chains. PRM structures are generated using the $add\_edge$, $delete\_edge$ and $reverse\_edge$ operators and aggregation functions if needed (cf. Section~\ref{PRM}). As for score function, they used a relational extension of the Bayesian Dirichlet (BD)~\cite{cooper92} score expressed as follows:

\begin{small}
\begin{multline}
\mathcal RBD_{score} = \sum_{i}\sum_{A\in \mathcal A(X_{i})}\sum_{u \in V(Pa(X_{i}.A))}log [DM(\{C_{X_{i}.A}[v,u]\},\{\alpha_{X_{i}.A}[v,u]\})]\\
-\sum_{i}\sum_{A\in \mathcal A(X_{i})}\sum_{u \in V(Pa(X_{i}.A))} length_{\mathcal K}(X_{i}.A,Pa(X_{i}.A))
\end{multline}
\end{small}

Where 
\[DM(\{C_{X_{i}.A}[v,u]\},\{\alpha_{X_{i}.A}[v,u]\})=\frac{\Gamma(\sum_{v}\alpha[v])}{\Gamma(\sum_{v}(\alpha[v]+C[v]))}\prod_{v}\frac{\Gamma(\alpha[v]+C[v])}{\Gamma(\alpha[v])},
\]
and
\[\Gamma(x)=\int^{\infty}_{0}t^{x-1}e^{-t}dt
\]
is the Gamma function.\\

As for standard BNs, evaluating the effectiveness of the proposed approaches is needed. However, neither relational benchmarks nor general random generation process are available.\\

\textbf{Random probabilistic relational models generation} has to be established in order to evaluate proposed learning approaches in a common framework.~\cite{Maier10} used a predefined schema and have only generated a number of dependencies varying from 5 to 15 and the conditional probability tables for attributes from a Dirichlet distribution. In~\cite{Maier132}, the authors have generated relational synthetic data to perform experimentation. Their generation process is based only on a particular family of relational schemas, with $N$ classes (nodes) and $N-1$ referential constraints (edges). Referential constraints are then expressed using relationship classes. This gives rise to a final relational schema containing $2N-1$ relations whereas in real world cases, relational schemas may have more than $N-1$ referential constraints. If the schema is fully connected (as described in~\cite{corrMaier}), it will have a tree structure. Torti et al.~\cite{Torti10} proposed a slightly different representation of PRMs, developed in the basis of the object-oriented framework and expert knowledge. Their main issue is probabilistic inference rather than learning. In their experimental studies~\cite{Wuillemin12}, they have randomly generated PRMs using the layer pattern. The use of this architecture pattern imposes a particular order when searching for connections between classes, generating reference slots of the relational schema and also when creating the relational skeleton. No indication has been made about the generation of probabilistic dependencies between attributes. In addition, they are  interested neither in populating a relational database nor in communicating with a database management system.

\section{PRM Benchmark Generation}
\label{cont}
Due to the lack of famous PRMs in the literature, this paper proposes a synthetic approach to randomly generate probabilistic relational models from scratch and to randomly instantiate them and populate relational databases. To the best of our knowledge, this has not yet been addressed.

\subsection{Principle}
As we are working with a relational variety of Bayesian networks, our generation process is inspired from classical methods of generation of random BNs while respecting the relational domain representation. 

The overall process is outlined in Algorithm~\ref{Overallalgo}. Roughly, the proposed generation process is divided into three main steps:
\begin{itemize}
	\item The first step generates a random PRM. For this, a relational schema is generated using $Generate\_Relational\_Schema$ function(Section~\ref{schemageneration}). Then, a graph dependency structure is generated using $Generate\_\\Dependency\_Structure$ and $Determine\_Slot\_Chains$ functions (Section~\ref{PRMgeneration}). And finally,  conditional probability tables are generated by the $Generate\_CPD$ function in the same way than Bayesian networks (cf. Section~\ref{BN}).
	\item The second step instantiates the model generated in the first step. First, a relational skeleton is generated using  $Generate\_Relational\_Skeleton$ function (Section~\ref{Instancegeneration}). Then, using $Create\_GBN$ function, a ground Bayesian Network is generated from both the generated PRM and the generated relational skeleton. 
	\item The third step presents the $Sampling$ function. It involves database instance population and can be performed using a standard sampling method over the $GBN$ (Section~\ref{DBpop}).
\end{itemize}

\begin{algorithm}[!t]
\KwIn{$N:$ the number of relations, $\mathcal K_{max}:$ The maximum slot chain length allowed}
\KwOut{$\Pi:<\mathcal R, \mathcal S, CPD>$, $DB\_Instance$}
\BlankLine
\Begin{
\textit{Step 1: Generate the PRM}
\BlankLine
$\Pi.\mathcal R \leftarrow Generate\_Relational\_Schema(N)$
\BlankLine
$\Pi.\mathcal S \leftarrow Generate\_Dependency\_Structure(\Pi.\mathcal R)$
\BlankLine
$\Pi.\mathcal S \leftarrow Determine\_Slot\_Chains(\Pi.\mathcal R, \Pi.\mathcal S, \mathcal K_{max})$
\BlankLine
$\Pi.CPD \leftarrow Generate\_CPD(\Pi.\mathcal S)$
\BlankLine
\textit{Step 2: Instantiate the PRM}
\BlankLine
$\sigma_{r} \leftarrow Generate\_Relational\_Skeleton(\Pi.\mathcal R)$
\BlankLine
$GBN \leftarrow Create\_GBN(\Pi, \sigma_{r})$
\BlankLine
\textit{Step 3: Database population}
\BlankLine
$DB\_Instance \leftarrow Sampling(GBN) $
}
\caption{Generate\_Random\_PRM-DB}
\label{Overallalgo}
\end{algorithm}

\subsection{Generation of a random relational schema}
\label{schemageneration}
The relational schema generation process is depicted in Algorithm~\ref{algogrs}. Our aim is to generate a relational schema with a given number of classes such that it does not contain any referential cycles and also respects the relational model definition presented in section~\ref{RBD}. We apply concepts from the graph theory for random schema generation. We associate this issue to a DAG structure generation process, where nodes represent relations and edges represent referential constraints definition. $X_{i}\rightarrow X_{j}$ means that $X_{i}$ is the referencing relation and $X_{j}$ is the referenced one. Besides, we aim to construct schemas where $\forall {X_{i},X_{i}} \in \mathcal X$ there exists a referential path from $X_{i}$ to $X_{j}$. This assumption allows to browse all classes in order to discover probabilistic dependencies later and it is traduced by searching DAG structures containing a single connected component (i.e., connected DAG). 
\begin{algorithm}[!t]
\KwIn{$N:$ the required number of classes}
\KwOut{$\mathcal R:$ The generated relational schema}
\BlankLine
\Begin{
\Repeat{
$\mathcal G$ is a connected DAG
}{$\mathcal G \leftarrow Generate\_DAG(Policy)$}
\BlankLine
\For{each relation $X_{i} \in \mathcal R$}{
$\mathcal Pk\_X_{i}\leftarrow Generate\_Primary\_Key(X_{i})$
\BlankLine
$\mathcal A(X_{i})\leftarrow Generate\_Attributes(Policy)$
\BlankLine
$\mathcal V(X_{i}.A)\leftarrow Generate\_States(Policy)$
}
\BlankLine
\For{each $n_{i} \rightarrow n_{j} \in \mathcal G$}{
\BlankLine
$Fk\_X_{i}\leftarrow Generate\_Foreign\_Key(X_{i},X_{j},Pk\_X_{j})$
}
}
\caption{$Generate\_Relational\_Schema$}
\label{algogrs}
\end{algorithm}

Having a fixed number of relations $N$, the $Generate\_DAG$ function constructs a DAG structure $\mathcal G$ with $N$ nodes, where each node $n_{i} \in \mathcal G$ corresponds to a relation $X_{i} \in \mathcal R$ following various possible implementation policies (cf. Section~\ref{policies}). For each class, we generate a primary key attribute using $Generate\_Primary\_Key$ function. Then, we randomly generate the number of attributes and their associated domains using  $Generate\_Attributes$ and $Generate\_States$ functions respectively. Note that the generated domains do not take into account possible probabilistic dependencies between attributes. For each $n_{i}\rightarrow n_{j} \in \mathcal G$, we generate a foreign key attribute in $X_{i}$ using the $Generate\_Foreign\_Key$ function. Foreign key generation is limited to one attribute as foreign keys reference simple primary keys (i.e., primary keys generated from only one attribute).\\

\subsection{Generation of a random PRM}
\label{PRMgeneration}
Relational schemas are not sufficient to generate databases when the attributes are not independent. We need to randomly generate probabilistic dependencies between the attributes of the classes in the schema. These dependencies have to provide the DAG of the dependency structure $\mathcal S$ and a set of CPDs which define a PRM (cf. Definition~\ref{def3}).

We especially focus on the random generation of the dependency structure. Once this latter is identified, conditional probability distributions may be sampled in a similar way as standard BNs parameter generation.

The dependency structure $\mathcal S$ should be a DAG to guarantee that each generated ground network is also a DAG~\cite{getoor02}. $\mathcal S$ has the specificity that one descriptive attribute may be connected to another with different possible slot chains. Theoretically, the number of slot chains may be infinite. In practice a user-defined maximum slot chain length $K_{max}$, is specified to identify the horizon of all possible slot chains. In addition, the $K_{max}$ value should be at least equal to $N-1$ in order to not neglect potential dependencies between attributes of classes connected via a long path. Each edge in the DAG has to be annotated to express from which slot chain this dependency is detected. We add dependencies following two steps. First we add oriented edges to the dependency structure while keeping a DAG structure. Then we identify the variable from which the dependency has been drawn by a random choice of a legal slot chain related to this dependency.

\subsubsection{Constructing the DAG structure} The DAG structure identification is presented in Algorithm~\ref{depStructAlgo}. The idea here is to find, for each node $X.A$, a set of parents from the same class or from further classes while promoting intra-class dependencies in order to control the final model complexity as discussed in~\cite{getoor02}. This condition promotes the discovery of intra-class dependencies or those coming from short slot chains. The longer the slot chain, the lower is the chance of finding a probabilistic dependency through the slot chain. To follow this condition, having $N$ classes, we propose to construct $N$ separated sub-DAGs, each of which is built over attributes of its corresponding class using the $Generate\_Sub\_DAG$ function. Then, we construct a super-DAG over all the previously constructed sub-DAGs. At this stage, the super-DAG contains $N$ disconnected components: The idea is to add inter-class dependencies in such a manner that we connect these disconnected components while keeping a global DAG structure.

To add inter-class dependencies, we constrain the choice of adding dependencies among only variables that do not belong to the same class. For an attribute $X.A$, the $Generate\_Super\_DAG$ function chooses randomly an attribute $Y.B$, where $X \neq Y$, then verifies whether the super-DAG structure augmented by a new dependency from $X.A$ to $Y.B$ remains a DAG. If so, it keeps the dependency otherwise it rejects it and searches for a new one. The policies that are used are discussed in Section~\ref{policies}.

\begin{algorithm}[!t]
\KwIn{$\mathcal R:$ The relational schema}
\KwOut{$\mathcal S:$ The generated relational dependency structure}
\BlankLine
\Begin{
\For{ each class $X_{i} \in \mathcal R$}{
$\mathcal G_{i} \leftarrow Generate\_Sub\_DAG(Policy)$ 
}
\BlankLine
$\mathcal S \leftarrow \bigcup \mathcal G_{i}$
\BlankLine
$\mathcal S \leftarrow Generate\_Super\_DAG(Policy)$
}
\caption{Generate\_Dependency\_Structure}
\label{depStructAlgo}
\end{algorithm}

\begin{algorithm}[!t]
\KwIn{$\mathcal R:$ The relational schema, $\mathcal S:$ The dependency structure, $K_{max}:$ The maximum slot chain length}
\KwOut{$\mathcal S:$ The generated relational dependency structure with generated slot chains}
\BlankLine
\Begin{
$K_{max} \leftarrow max(K_{max}, card(\mathcal X_{\mathcal R})-1)$
\BlankLine
\For{each $X.A \rightarrow Y.B \in \mathcal S$}{
$Pot\_Slot\_Chains\_List \leftarrow Generate\_Potential\_Slot\_chains(X, Y, \mathcal R, K_{max})$
\BlankLine
\For{each $slot\_Chain \in Pot\_Slot\_Chains\_List$}{
\BlankLine
$l \leftarrow length(slot\_Chain )$
\BlankLine
$W[i]\leftarrow \exp^{\frac{-l}{nb\_Occ(l,Pot\_Slot\_Chains\_List)}}$
}
\BlankLine
$Slot\_Chain^{*}\leftarrow Draw(Pot\_Slot\_Chains\_List, W)$
\BlankLine
\If{$Needs\_Aggregator(Slot\_Chain^{*})$}{$\gamma \leftarrow Random\_Choice\_Agg(list\_Aggregators)$}
\BlankLine
\eIf{$Slot\_Chain^{*}=0$}{$\mathcal S.Pa( X.A)\leftarrow \mathcal S.Pa( X.A)\cup Y.B$ \textit{\% here $X=Y$}}{$\mathcal S.Pa(X.A)\leftarrow \mathcal S.Pa(X.A)\cup \gamma(Y.Slot\_Chain^{*}.B)$}
}
}
\caption{Determine\_Slot\_Chains}
\label{SCAlgo}
\end{algorithm}
\subsubsection{Determining slot chains} 
During this step, we have to take into consideration that one variable may be reached through different slot chains and the dependency between two descriptive attributes will depend on the chosen one. Following~\cite{getoor02}, the generation process has to give more priority to shorter slot chains for selection. Consequently, we have used the penalization term discussed in~\cite{getoor02}. Longer indirect slot chains are penalized by having the probability of occurrence of a probabilistic dependency from a slot chain length $l$ inversely proportional to $\exp^{l}$.

Having a dependency $X.A\rightarrow Y.B$ between two descriptive attributes $X.A$ and $Y.B$, we start by generating the list of all possible slot chains ($Pot\_Slot\_Chains\_List$) of $length\leq K_{max}$ from which $X$ can reach $Y$ in the relational schema using the $Generate\_Potential\_Slot\_chains$ function. Then, we create a vector $W$ of the probability of occurrence for each of the slot chains found, with $\log (W[i]) \propto \frac{-l}{nb\_Occ(l,Pot\_Slot\_Chains\_List)}$, where $l$ is the slot chain length and $nb\_Occ$ is the number of slot chains of length $l \in Pot\_Slot\_Chains\_List$. This value will rapidly decrease when the value of $l$ increases, which allows to reduce the probability of selecting long slot chains. We then sample a slot chain from $Pot\_Slot\_Chains\_List$ following $W$ using the $Draw$ function. If the chosen slot chain implies an aggregator, then we choose it randomly from the list of existing ones using the $Random\_Choice\_Agg$ function. The slot chain determination is depicted in Algorithm~\ref{SCAlgo}.

\paragraph{Simplifying slot chains.} While finding slot chains,  duplicate slot chains might be encountered. By 'duplicate', we mean the slot chains which produce the same result. For example, in the schema of figure \ref{subfig1}, $Vote.User$ and $Vote.User.User^{-1}.User$ are equivalent because traversing through the slot chains, we obtain the same set of User objects. Similarly, $Vote.Movie^{-1}.Movie$ is the same as an empty slot chain because this slot chain results in the target Movie object. When such duplicates are found, we pick the shorter one to avoid redundant, unnecessary computations. This is an improvement to our previous work~\cite{mouna14,BenIshakIDA}, where simplification of slot chains had not been considered. We apply the following rule to simplify slot chains.

A slot chain is represented as a sequence of reference slots and inverse slots as $\rho_1.\rho_2. \dots .\rho_{n-1}.\rho_n$. If $\rho_{n-1}$ is an inverse slot and $\rho_n^{-1} = \rho_{n-1}$, then the slot chain can be simplified by eliminating the last two slots. So, the simplified slot chain would, then, be $\rho_1.\rho_2. \dots .\rho_{n-3}.\rho_{n-2}$. This can be done repetitively until no simplification is possible.\\

\subsection{GBN generation}
\label{Instancegeneration}

The generated schema together with the added probabilistic dependencies and generated parameters results in a probabilistic relational model. To instantiate this latter, we need to generate a relational skeleton. The GBN is, then, fully determined with this relational skeleton and the CPDs already present at the meta-level.

A relational skeleton can be imagined as a DAG where nodes are objects of different classes present in the associated relational schema and edges are directed from one object to another conforming to the reference slots present in the relational schema. This graph is, in fact, a special case of $k$-partite graph\footnote{A $k$-partite graph is a graph whose vertices can be partitioned into $k$ disjoint sets so that there is no edge between any two vertices within the same set.} of definition \ref{def:skeleton-graph}.

\begin{definition}{Relational skeleton as a $k$-partite graph}{\label{def:skeleton-graph}}
	
	A relational skeleton is a special case of $k$-partite graph, $G_k = (V_k, E_k)$, with the following properties:
	
	\begin{enumerate}
		\item The graph is acyclic, 
		\item All edges are directed (an edge $u \rightarrow v$ indicates that the object $u$ refers to $v$, i.e., $u$ has a foreign key which refers to the primary key of $v$),
		\item Edges between two different types of objects are always oriented in the same direction, i.e. for all edges ($u$ --- $v$) between objects of $U$ and $V$ where $u \in U$, and $v \in V$, the direction of all edges must be either $u \rightarrow v$ or $u \leftarrow v$ and not both
		\item For all edges $u \rightarrow v$ between the objects of $U$ and $V$, out degree of $u = 1$ but indegree of $v$ can be greater than 1.
		
	\end{enumerate}
\end{definition}

In this regard, relational skeleton generation process can be considered as a problem of generating objects and assigning links (or foreign keys) between them such that the resulting graph is a $k$-partite graph of definition \ref{def:skeleton-graph}. In our previous work \cite{mouna14,BenIshakIDA}, we presented an algorithm to generate relational skeleton, where it generates nearly same number of objects of each class and iteratively adds random edges between objects of a pair of classes such that the direction of the edges conform to the underlying schema. This approach, in fact, does not create realistic skeleton because in real world, relational skeleton tends to be scale-free, i.e., degree of the vertices of the graph follows power law. Hence, in real datasets, the number of objects for classes with foreign keys tend to be very high compared to that for classes which do not have foreign keys and are referenced by other classes. Thus, we took a different approach to generate relational skeleton. Our new and improved approach to generating such $k$-partite graph is presented as Algorithm \ref{algo:relation-skeleton}. We adapt \cite{bollobas2003directed}'s directed scale-free graph generation algorithm for our special $k$-partite graph and use Chinese Restaurant Process\cite{pitman2002combinatorial} to apply preferential attachment.

\begin{algorithm}[!t]
\KwIn{Relational Schema as a DAG, $\mathcal{G} = (V_g, E_g)$; Total number of objects in the resulting skeleton, $N_{total}$; Scalar parameter for CRP, $\alpha$}
\KwOut{A relational skeleton, $\mathcal{I}=(V, E)$}
\BlankLine
\Begin{
		
		\For {$node \in V_g $}{
		$N(node) \gets 0$ \textit{\%Total number of objects of each type generated so far}
		}
		\BlankLine
		$V \gets \{\}$	\textit{\%Set of objects}
		\BlankLine
		$E \gets \{\}$	\textit{\%Set of directed edges between objects}
		\BlankLine
		$m \gets $ Number of nodes without any parents in $\mathcal{G}$ \textit{\%number of roots}
		\BlankLine
		\If {$m > 1$}{
		Divide $\mathcal{G}$ into $m$ subgraphs such that each subgraph contains a root and all of its descendants.
		}
		\Repeat{$n >= N_{total}$}{
		\eIf {$m > 1$} {
				$g \gets$ one of the $m$ subgraphs picked randomly}
				{	
				$g \gets \mathcal{G}$ \textit{\%i.e., if $\mathcal{G}$ has only one root}
			}
			\BlankLine
			$obj_{root} \gets $ A new object of the root of $g$ 
			\BlankLine
			$n_{root} \gets n_{root}+1$
			\BlankLine
			$children \gets$ Children of the root in $g$
			\BlankLine
			$((V', E'), N') \gets $ Generate\_SubSkeleton($obj_{root}, g, children, N, \alpha$) \textit{\%Perform depth-first search over $g$ and add edges recursively}
			\BlankLine
			$V \gets V \cup V'$
			\BlankLine
			$E \gets E \cup E'$
			\BlankLine
			$N \gets N'$		\textit{\%Update the set of number of generated objects of each type}
			\BlankLine
			$n \gets cardinality(V)$ 	\textit{\%Total number of objects generated so far.}
		}
		\BlankLine
		$\mathcal{I} \gets (V, E)$
}
\caption{Generate\_Relational\_Skeleton}
\label{algo:relation-skeleton}
\end{algorithm}

The basic idea here is to iteratively generate an object of a class with no parents in the relational schema DAG and then recursively add an edge from this object to objects of its children classes. This process is essentially a depth first search (DFS), where we begin by generating an object of the root node of the graph and then at each encounter of a node in DFS, we add an edge from the object of the parent node to either a new or an existing object of the encountered node. The object of the parent node gets connected to a new object with probability $p = \alpha/(n_p-1+\alpha)$, where $n_p$  is the total of objects of the parent node generated so far, and $\alpha$ is a scalar parameter for the process. When it gets attached to an existing object, an object of the correct type is picked from the set of existing objects with probability $n_k/(n_p-1 + \alpha)$, where $n_k$ is the indegree of the object to be selected and $n$ is the total number of objects generated so far. Thus, as the skeleton graph grows, probability of getting connected to new objects will decrease and the objects with higher indegree will be preferred for adding new edges. At each iteration, a DFS is performed starting from one of the nodes without parents in the relational schema DAG. Thus, if there is only one node that does not have any parent, then each iteration will visit all classes in the relational schema resulting in a complete set of objects and relations for all classes, otherwise only a subset of classes will be visited in each iteration. So, at the beginning of each iteration, one of the nodes without parents is picked randomly in the latter case. 
The iteration process is continued until the skeleton contains the required number of objects.\\

\begin{algorithm}[!t]
\KwIn{Parent object $obj_p$; Graph $g$; Parent node, $parent$; Children nodes, $children$; Set of the number of objects of each class generated so far, $N$; Scalar parameter $\alpha$}
\KwOut{Relational skeleton, $\mathcal{I}=(V, E)$; Set of the number of objects of each class generated so far, $N$}
\BlankLine
\Begin{
$V \gets \{obj_p\}$
\BlankLine
$E \gets \{\}$
\BlankLine
$n_p \gets N(parent)$ \textit{\%Total number of parents generated so far}
\BlankLine
\For {$C \in children$}{
			$n_c \gets N(C)$ \textit{\%Total number of the child $C$ generated so far}
			\BlankLine
			$p \gets \alpha / (n_p - 1 + \alpha)$
			\BlankLine
			$r \gets$ A random value between 0 and 1 \textit{\%$r \in [0, 1]$}
			\BlankLine
			\eIf {$r <= p$}{
				$obj_c \gets$ Create a new object of type $C$
				\BlankLine
				$N(C) \gets n_c + 1$
				\BlankLine
				$e_{pc} \gets (obj_p, obj_c)$ \textit{\%Add an edge from $obj_p$ to $obj_c$}
				\BlankLine
				$E \gets E \cup \{e_{pc}\}$
				\BlankLine
				$children_c \gets$ Children of $C$ in the graph $g$
				\BlankLine
				$((V', E'), N') \gets $ Generate\_SubSkeleton($obj_c, g, C, children_c, N, \alpha$)
				\BlankLine
				$V \gets V \cup V'$
				\BlankLine
				$E \gets E \cup E'$
				\BlankLine
				$N \gets N'$}{ 
				$obj_c \gets $ An existing object of type $C$ picked randomly with probability $n_k/(n_p-1)$ where $n_k$ = indegree of $obj_c$
				\BlankLine
				 $e_{pc} \gets (obj_p, obj_c)$ \textit{\%Add an edge from $obj_p$ to $obj_c$}
				\BlankLine
				$E \gets E \cup \{e_{pc}\}$
			}
}
\BlankLine
$\mathcal{I} \gets (V, E)$ 
}
\caption{Generate\_SubSkeleton}
\label{algo:gen-subskeleton}
\end{algorithm}

\subsection{Database population} 
\label{DBpop}
This process is equivalent to generating data from a Bayesian network. We can generate as many relational database instances as needed by sampling from the constructed GBN. The specificity of the generated tuples is that they are sampled not only from functional dependencies but also from probabilistic dependencies provided by the randomly generated PRM. \\

\section{Toy example}
\label{example}

In this section, we illustrate our proposal through a toy example.\\

\begin{figure}[!t]
	\centering
	\includegraphics[width=0.95\textwidth]{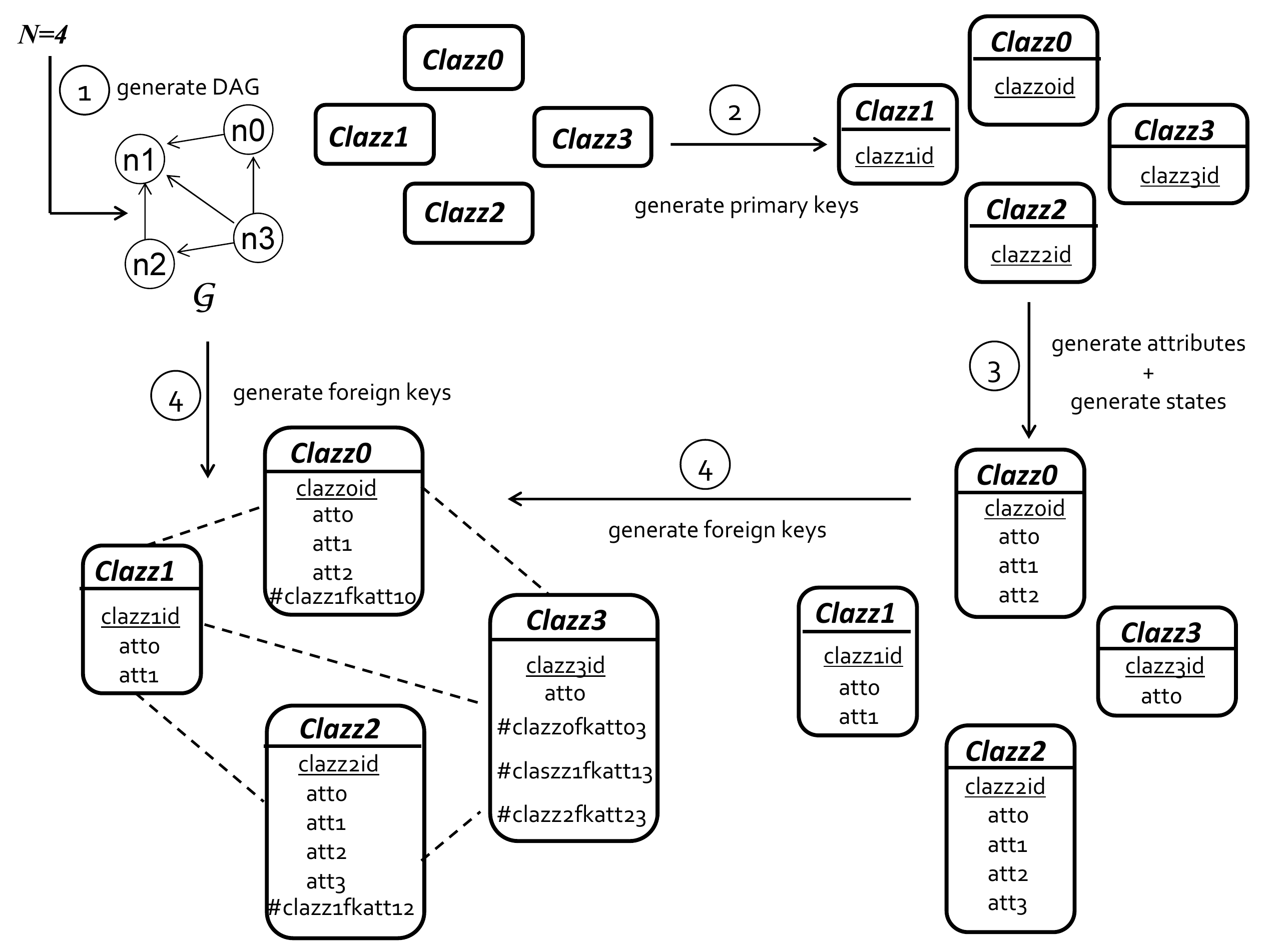}
	\caption{Relational schema generation steps}
	\label{fig3}
\end{figure}

\begin{figure}[!t]
	\centering
	\includegraphics[width=0.85\textwidth]{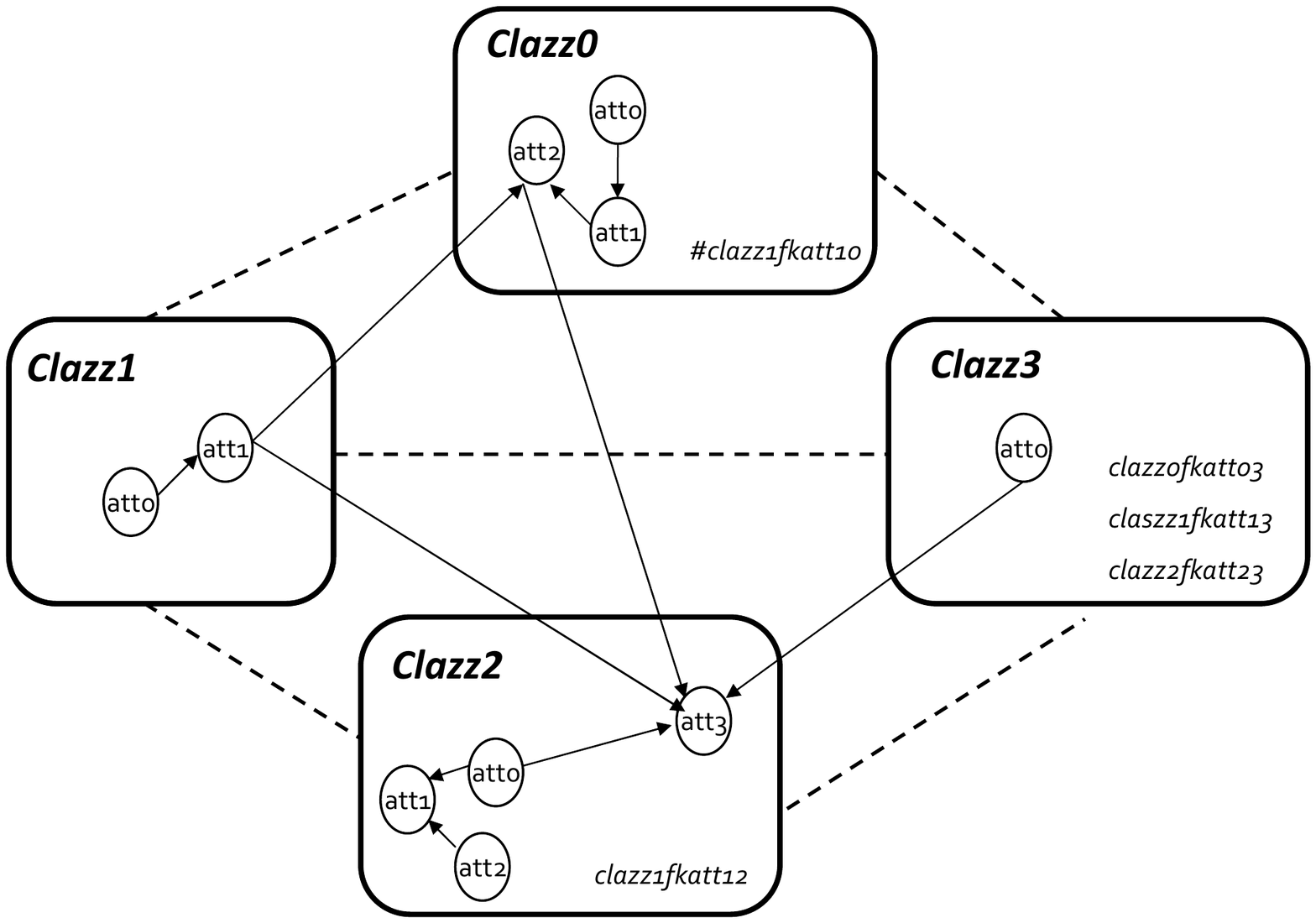}
	\caption{Graph dependency structure generation}
	\label{fig4}
\end{figure}

\begin{figure}[!t]
	\centering
	\includegraphics[width=0.85\textwidth]{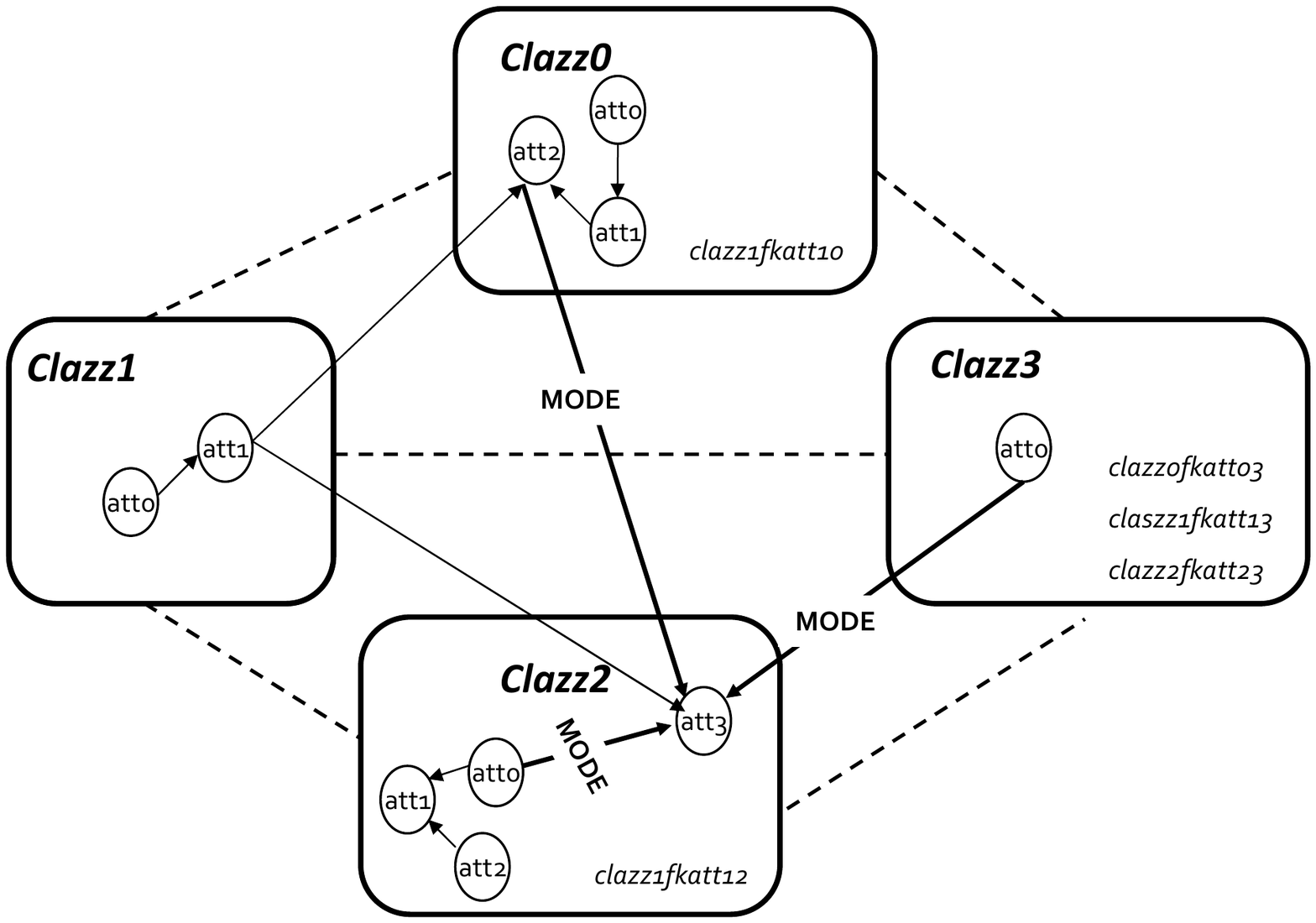}
	\caption{Example of a generated relational schema where the dotted lines represent referential constraints and the generated PRM dependency structure where the arrows represent probabilistic dependencies. Here, we have excluded slot chains to not overload the figure. Details about slot chains from which probabilistic dependencies have been detected are given in Paragraph \textit{PRM generation}.}
	\label{fig2}
\end{figure}

\textbf{Relational schema generation.} Figure~\ref{fig3} presents the result of running Algorithm~\ref{algogrs}, with $N=4$ classes. For each class, a primary key has been added ($clazz0id$, $clazz1id$, $clazz2id$ and $clazz3id$). Then a number of attributes has been generated randomly together with a set of possible states for each attribute using the policies described in Section~\ref{policies} (e.g., $clazz0$ has $3$ descriptive attributes $att0, att1$ and $att2$. $att0$ is a binary variable). Finally, foreign key attributes have been specified following the DAG structure of the graph $\mathcal G$ (e.g., $clazz2$ references class $clazz1$ using foreign key attribute $clazz1fkatt12$).\\

\textbf{PRM generation.} We recall that this process is performed in two steps: randomly generate the dependency structure $\mathcal S$ (Algorithm~\ref{depStructAlgo}), then randomly generate the conditional probability distributions which is similar to parameter generation of a standard BN. The random generation of the $\mathcal S$ is performed in two phases. We start by constructing the DAG structure, the result of this phase is in Figure~\ref{fig4}. Then, we fix a maximum slot chain length $K_{max}$ to randomly determine from which slot chain the dependency has been detected. We use $K_{max}=3$, the result of this phase gives rise to the graph dependency structure of Figure~\ref{fig2}. $\mathcal S$ contains $5$ intra-class and $5$ inter-class probabilistic dependencies. \\
Three of the inter-class dependencies have been generated from slot chains of length $1$:\\
$Clazz0.clazz1fkatt10.att1 \rightarrow Clazz0.att2$;\\
$MODE(Clazz2.clazz2fkatt23^{-1}.att0) \rightarrow Clazz2.att3$ and;\\
$Clazz2.clazz1fkatt12.att1\rightarrow Clazz2.att3$\\
One from slot chain of length $2$: \\
$MODE(Clazz2.clazz1fkatt12. clazz1fkatt12^{-1}.att0) \rightarrow Clazz2.att3$\\
One from slot chain of length $3$:\\
$MODE(Clazz2.clazz2fkatt23^{-1}.claszz1fkatt13. clazz1fkatt10^{-1}.att2) \rightarrow Clazz2.att3$\\

\begin{figure}[!t]
    \centering
	\begin{subfigure}[b]{0.3\textwidth} 
		\centering
		\includegraphics[width=\textwidth]{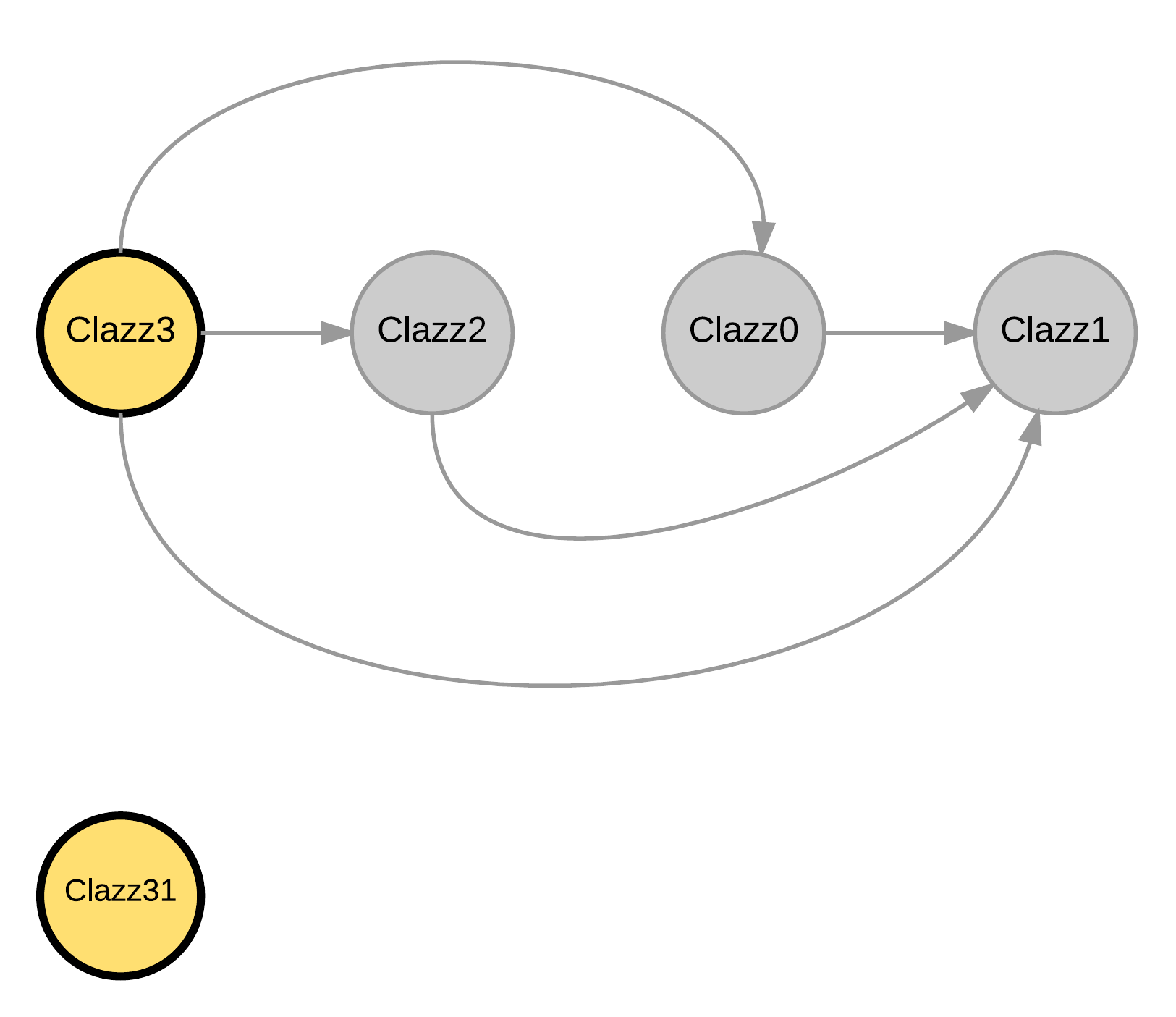}
		\caption{} 
		\label{fig:skeleton-dfs-1}
	\end{subfigure}
	\quad    
	\begin{subfigure}[b]{0.3\textwidth}
		\centering
		\includegraphics[width=\textwidth]{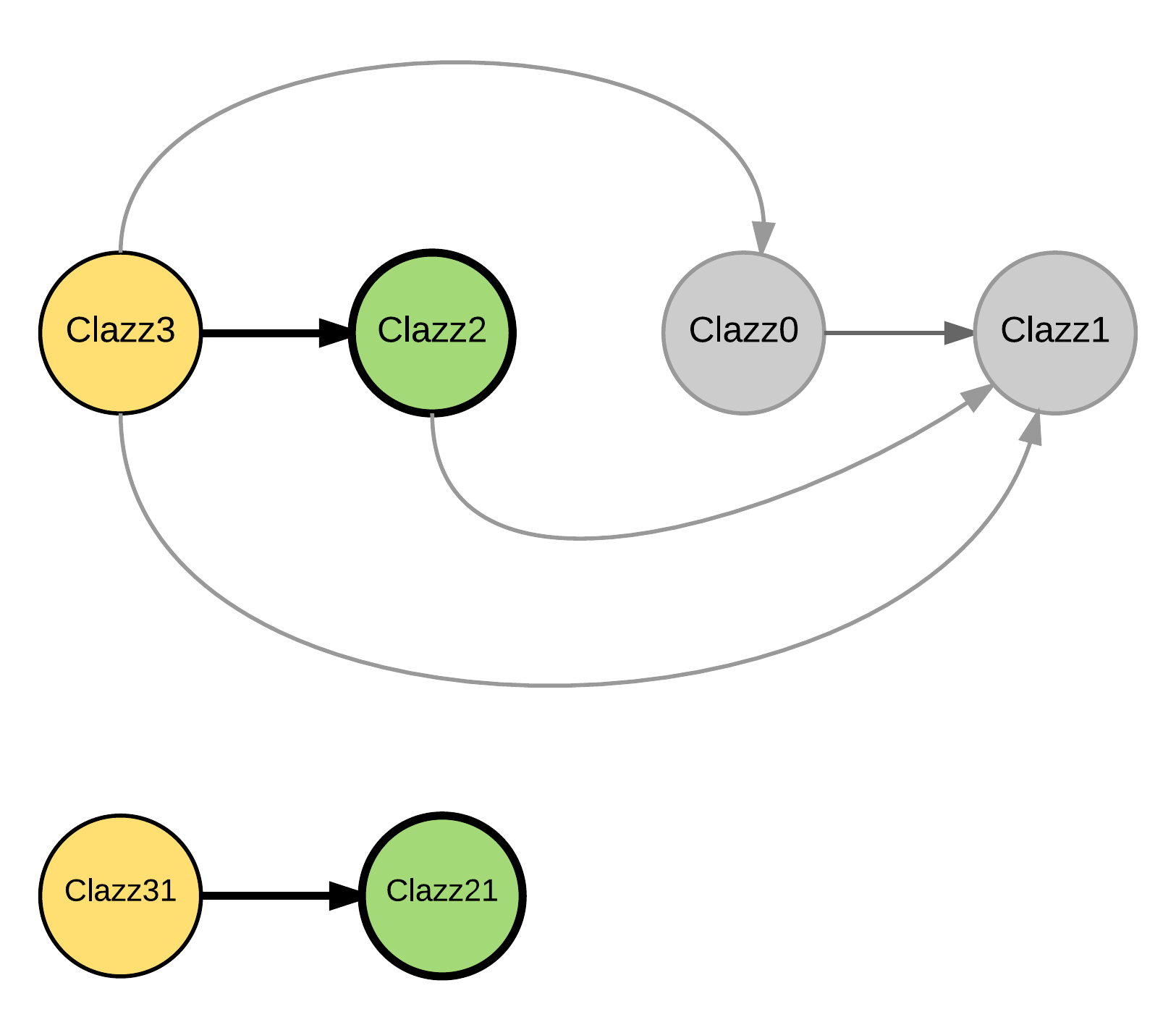}
		\caption{}
		\label{fig:skeleton-dfs-2}
	\end{subfigure}
	\quad    
	\begin{subfigure}[b]{0.3\textwidth}
		\centering
		\includegraphics[width=\textwidth]{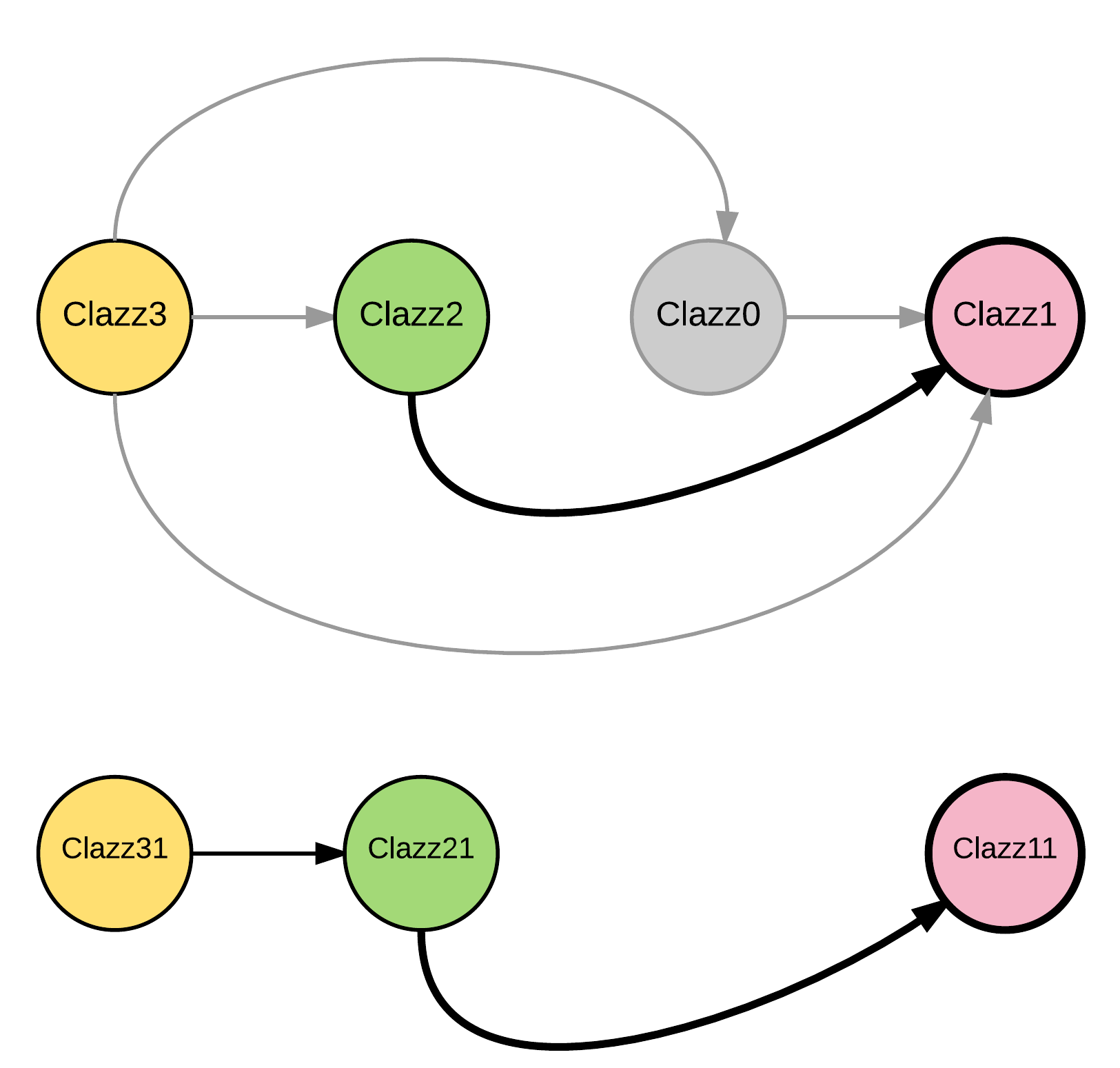}
		\caption{}
		\label{fig:skeleton-dfs-3}
	\end{subfigure}
	\hfill
	\hfill
	\begin{subfigure}[b]{0.3\textwidth} 
		\centering
		\includegraphics[width=\textwidth]{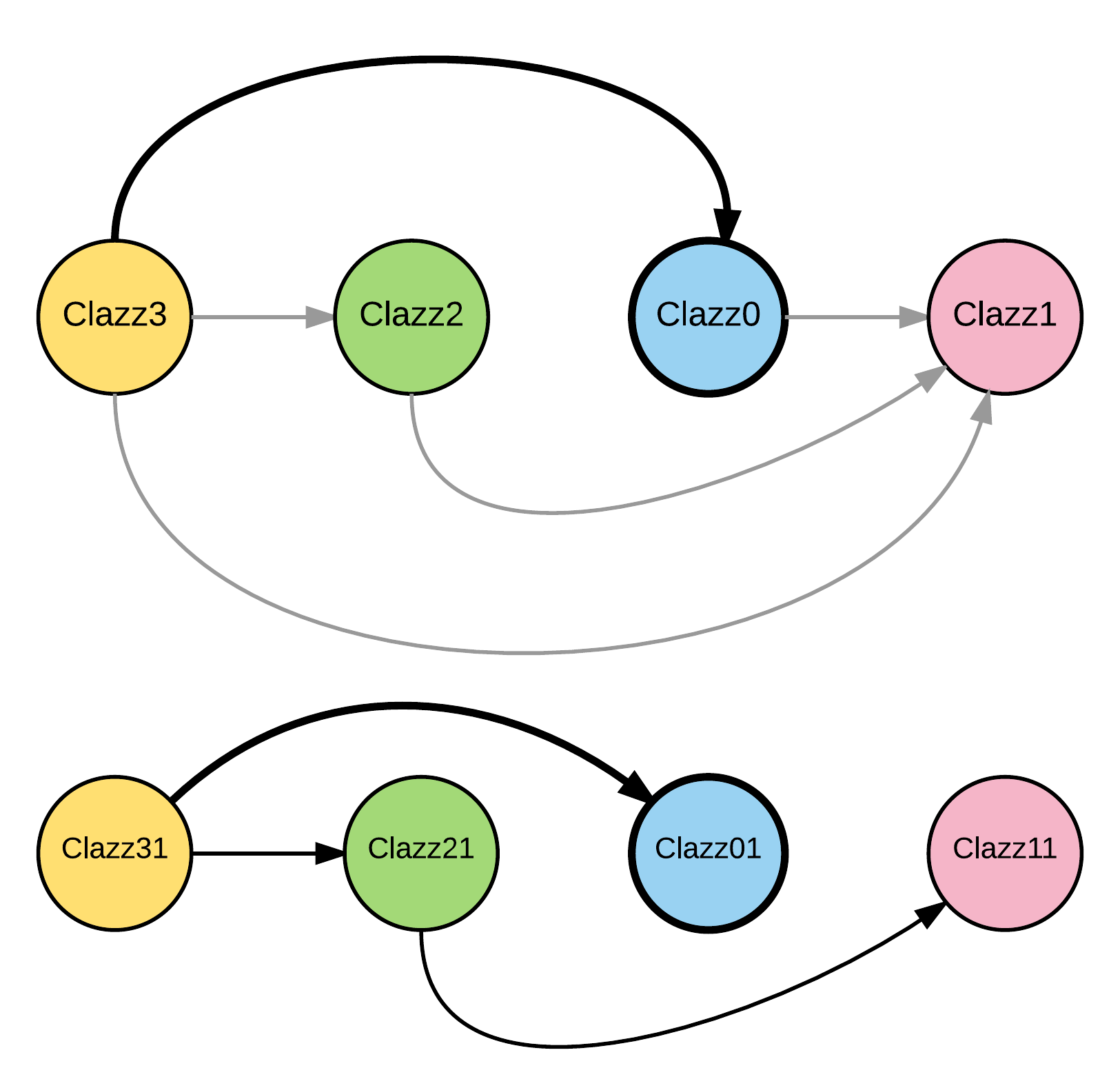}
		\caption{} 
		\label{fig:skeleton-dfs-4}
	\end{subfigure}
	\quad    
	\begin{subfigure}[b]{0.3\textwidth}
		\centering
		\includegraphics[width=\textwidth]{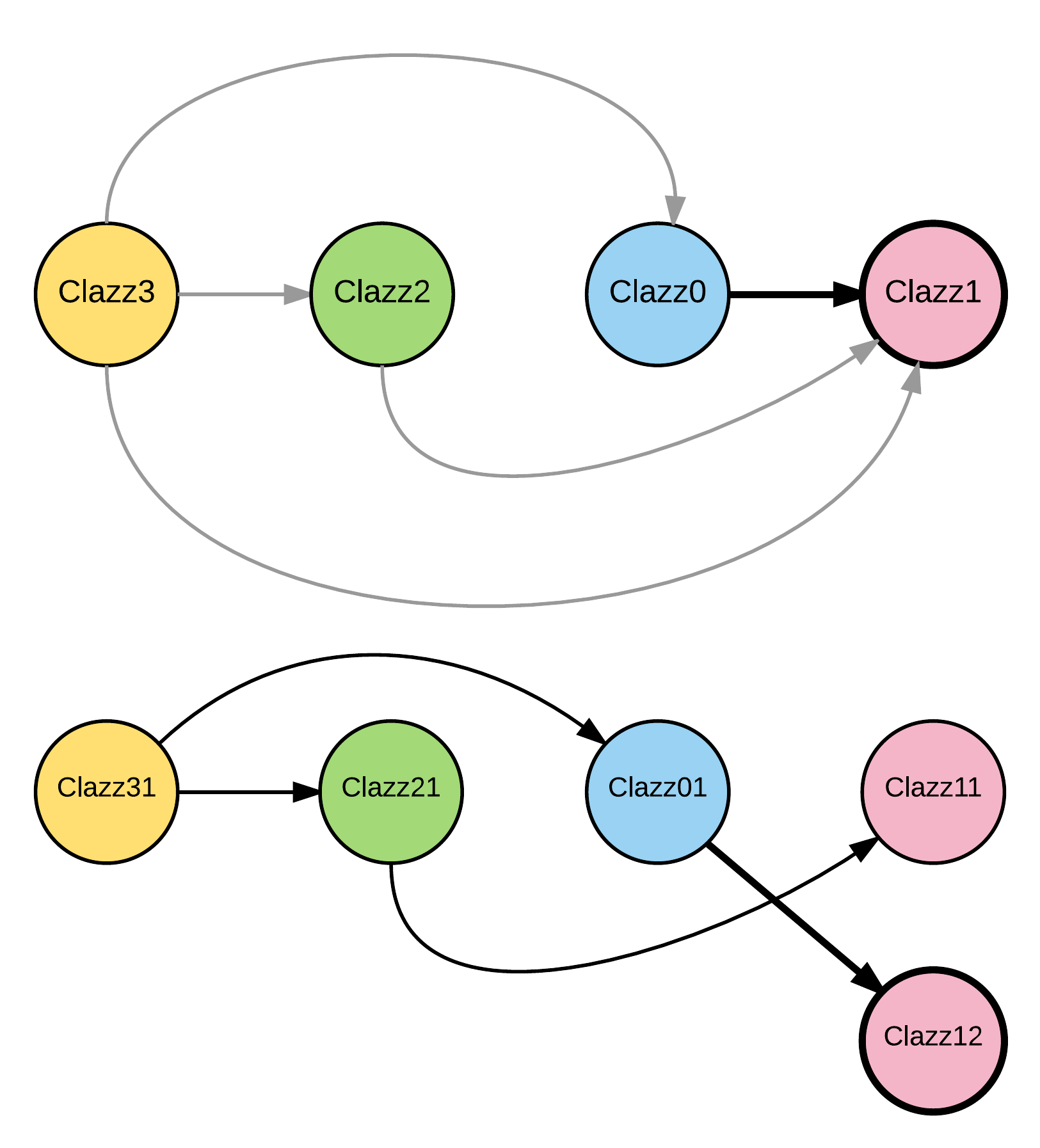}
		\caption{}
		\label{fig:skeleton-dfs-5}
	\end{subfigure}
	\quad    
	\begin{subfigure}[b]{0.3\textwidth}
		\centering
		\includegraphics[width=\textwidth]{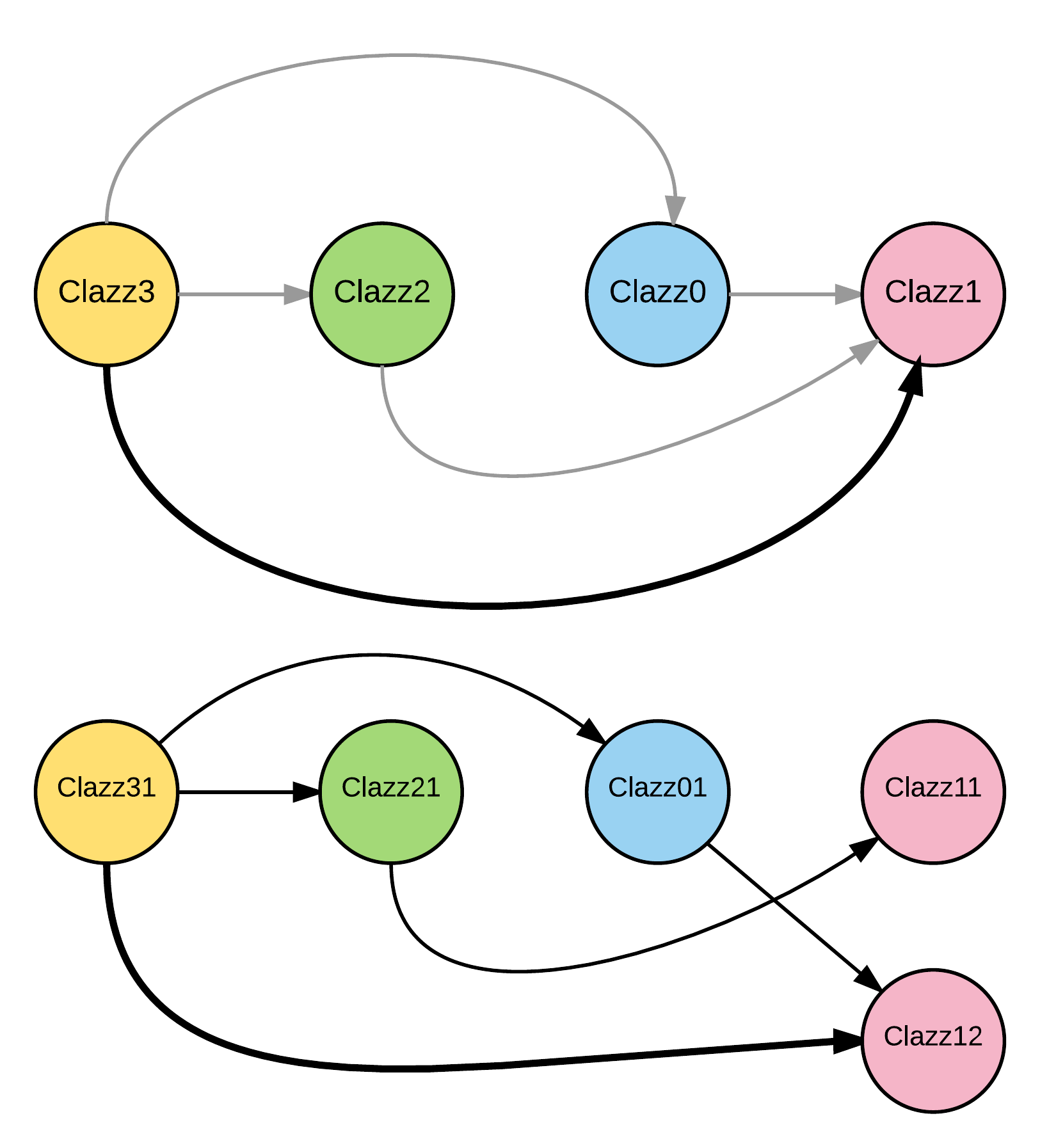}
		\caption{}
		\label{fig:skeleton-dfs-6}
	\end{subfigure}
	\caption{Generating objects while performing Depth First Search (DFS) on the relational schema. This figure shows one iteration of a DFS performed on the schema of figure \ref{fig3}. 
	In each of the sub-figures above, the upper graph is the concerned relational schema and the lower graph is the relational skeleton being generated. The node (and the edge) encountered at each step of the DFS is shown by thick lines. Colors are used to only as a visual aid to distinguish between objects of different classes and add no significant meaning to the process.
	(\ref{fig:skeleton-dfs-1}) We begin by creating a new object of the node `Class3' as it does not have any parent. 
	(\ref{fig:skeleton-dfs-2}) Then, we traverse to one of the children of this node in the schema (`Class2' here). As there is no object of this class so far, a new object will be created. 
	(\ref{fig:skeleton-dfs-3}) and (\ref{fig:skeleton-dfs-4}) Now, continuing the DFS, we encounter the node `Class1' (the child of `Class2'), and then `Class0' (a child of `Class3'). Like earlier, new objects of `Class1' and `Class0' will be generated. 
	(\ref{fig:skeleton-dfs-5}) As `Class0' has a child, we reach `Class1'. At this step, an object of `Class1' is already present. So the object `Class01' can either create a new object or get connected to `Class11'. In this example, it gets linked to a new object `Class12'.
	(\ref{fig:skeleton-dfs-6}) In the next step of the DFS, `Class1' is encountered again as it is a child of `Class3'. Here, `Class31' gets attached to an existing object of `Class1'. }
	\label{fig:skeleton-dfs}
	
\end{figure}

\begin{landscape}
\begin{figure}[!t]
    \centering
	\begin{subfigure}[b]{.7\textwidth}
		\centering
		\includegraphics[width=\textwidth]{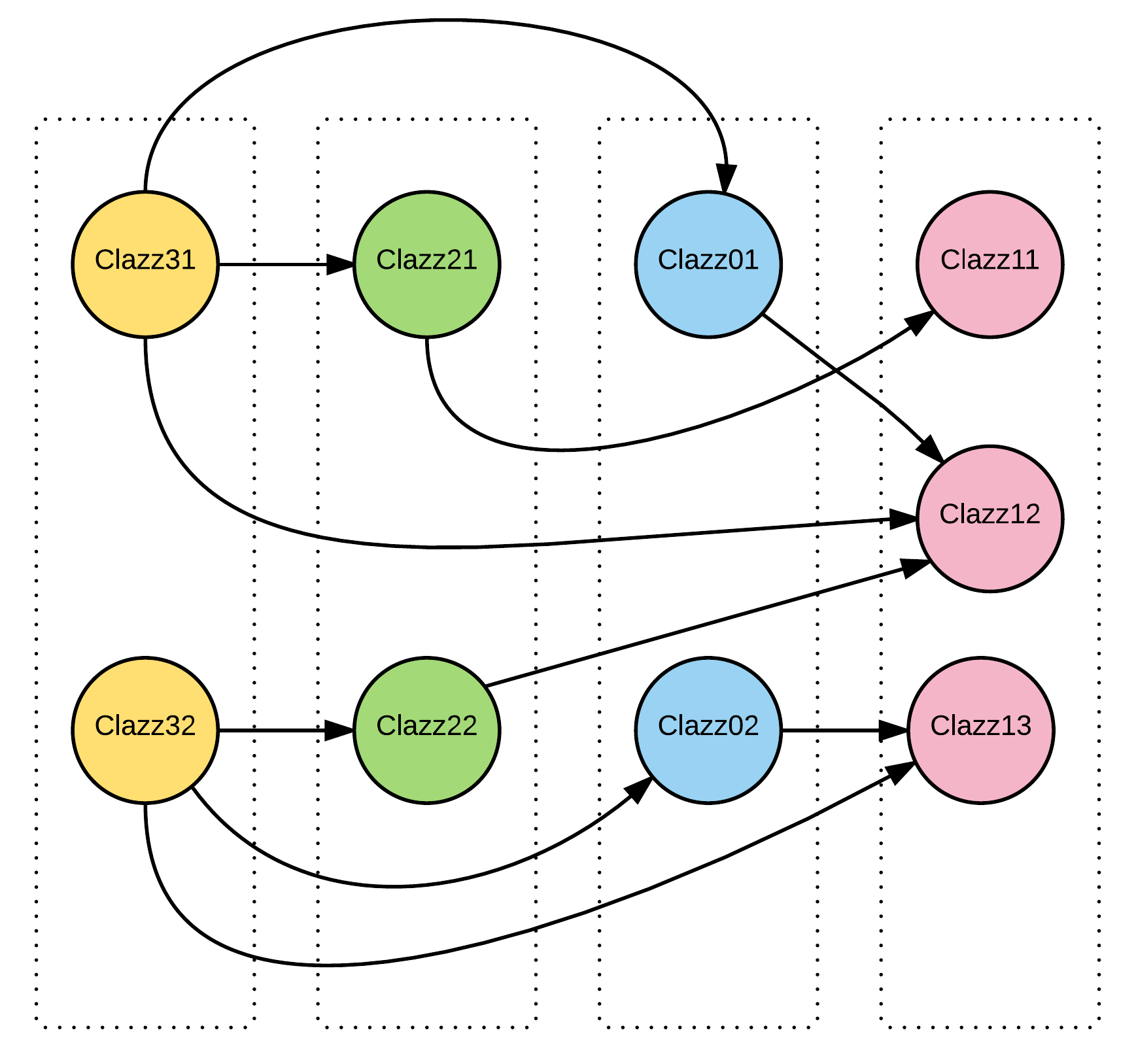}
		\caption{}
		\label{fig:skeleton-gen-iter2}
	\end{subfigure}
	\quad    
	\begin{subfigure}[b]{.7\textwidth}
		\centering
		\includegraphics[width=\textwidth]{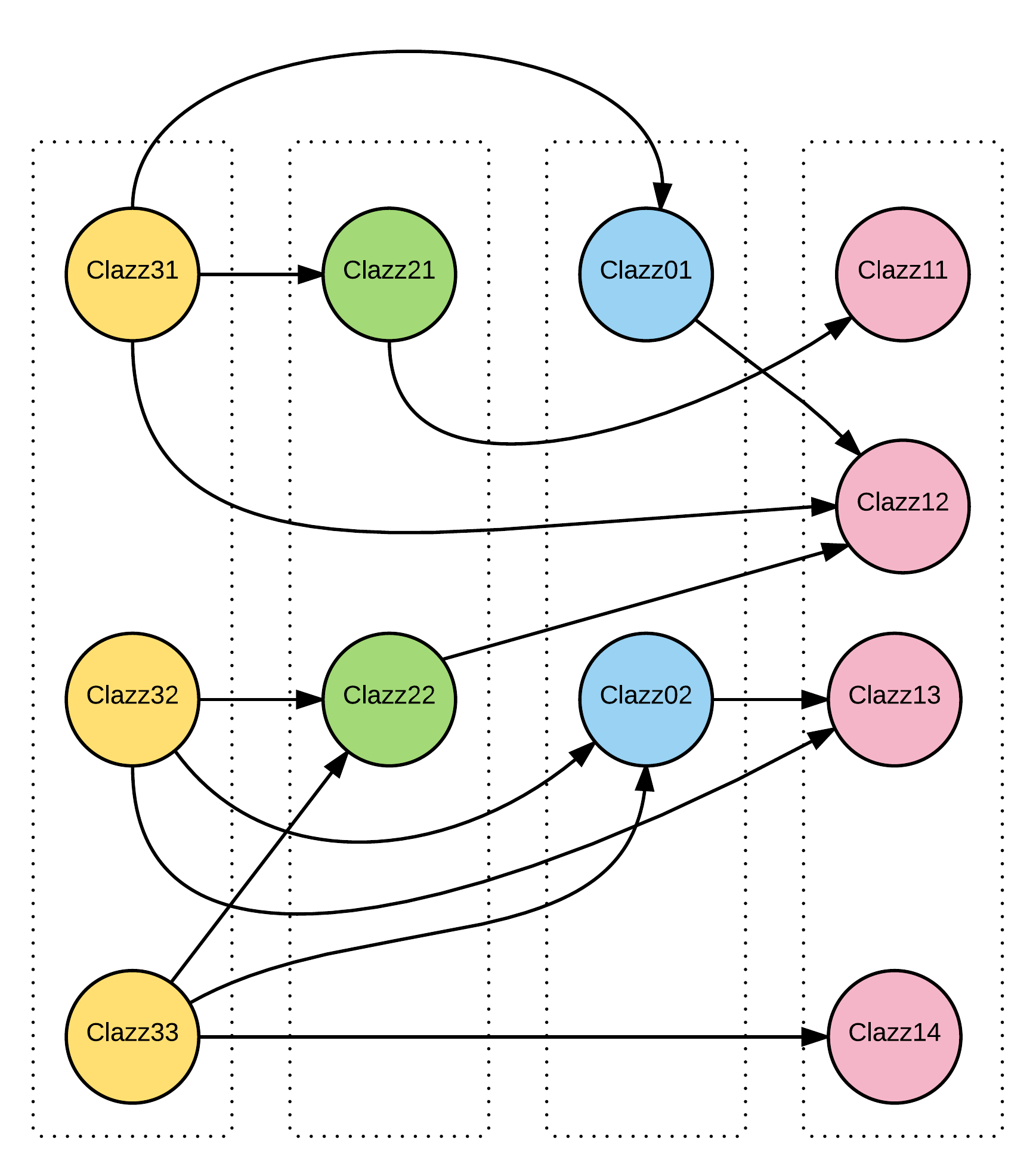}
		\caption{}
		\label{fig:skeleton-gen-iter3}
	\end{subfigure}
	\caption{Next two iterations of DFS on the relational schema of figure \ref{fig3} following the first iteration of figure \ref{fig:skeleton-dfs} to generate relational skeleton graph. At each iteration, a new object of `Class3' will always be generated as it does not have any parent. The object will then be linked to an existing object or a new one, and the same thing goes on for the new objects. Here, the skeleton after the first iteration has five objects. The second iteration creates four new objects, whereas the third iteration creates only two objects.}
	\label{fig:skeleton-generation}
\end{figure}

\end{landscape}

\begin{figure}[!t]
	\centering
	\includegraphics[width=0.8\textwidth]{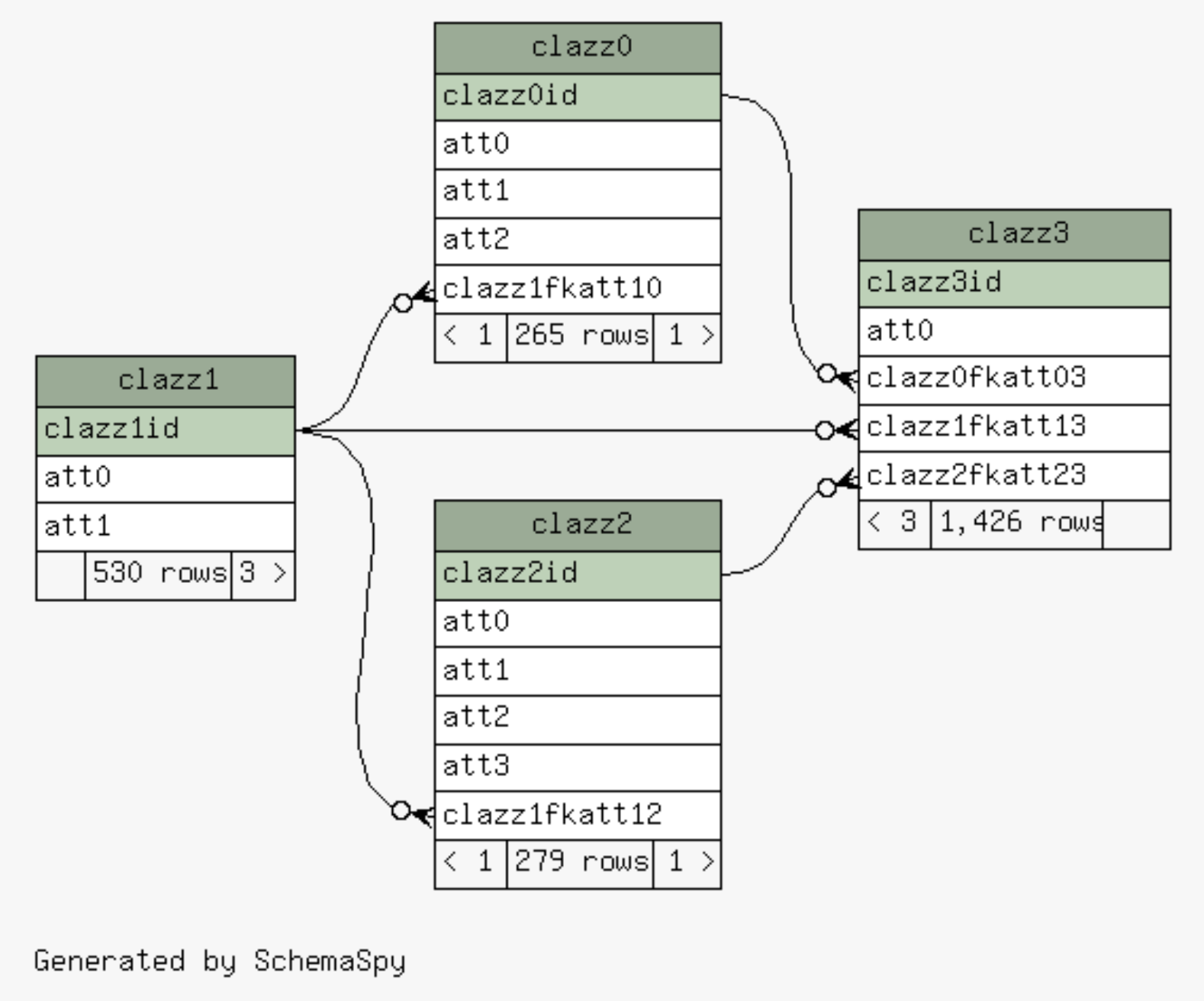}
	\caption{Schema diagram of our toy example showing the number of objects/rows in each class.}
	\label{fig:final-skeleton}
\end{figure}

\textbf{GBN creation.} Once the PRM is generated, we follow the two steps presented in Section~\ref{Instancegeneration} to create a GBN and to populate the DB instance. We create a relational skeleton for the relational schema by performing depth first search on the schema DAG~(cf.~Algorithm~\ref{algo:relation-skeleton}). The first three iterations of the DFS are shown in figures \ref{fig:skeleton-dfs} and \ref{fig:skeleton-generation}. As the schema has only one node without any parent (i.e., a class without any foreign key), one complete DFS returns a set of objects of each class as shown in figure \ref{fig:skeleton-dfs}. At each iteration, we obtain different number of objects. As we can see in figure \ref{fig:skeleton-generation}, the first iteration created five objects whereas the second and third iteration resulted in four and two objects respectively. We continue the iteration until we obtain the required number of objects in the skeleton. We then instantiate the probabilistic model generated in the previous step with the generation skeleton to obtain a ground Bayesian network. Sampling this GBN enables us to populate values for all attributes of all objects in the relational skeleton. For this example, we generated a random dataset with ~2500 objects. The corresponding schema diagram is shown in figure \ref{fig:final-skeleton}, which also shows the number of objects of each class. The diagram is generated using SchemaSpy\footnote{http://schemaspy.sourceforge.net/}.  

\section{Implementation}
This section explains the implementation strategy of our generator, identifies the chosen policies and discusses the complexity of the algorithms.

\subsection{Software implementation}
The proposed algorithms have been implemented in PILGRIM\footnote{http://pilgrim.univ-nantes.fr/
} API, a software platform that our lab is actively developing to provide an efficient tool to deal with several probabilistic graphical models (e.g., BNs, Dynamic BNs, PRMs). Developed in C++, PILGRIM uses Boost graph library\footnote{http://www.boost.org/} to manage graphs, ProBT API\footnote{http://www.probayes.com/fr/Bayesian-Programming-Book/downloads/} to manipulate BNs objects and Database Template Library (DTL)\footnote{http://dtemplatelib.sourceforge.net/dtl\_introduction.htm} to communicate with databases. Currently, only PostgreSQL RDBMS is supported in this platform.

Besides the algorithms, we have also implemented serialization of PRMs. Because there is currently no formalization of PRMs, we propose an enhanced version of the XML syntax of the \textit{ProbModelXML} specification\footnote{http://www.cisiad.uned.es/techreports/ProbModelXML.pdf} to serialize our generated models. We have added new tags to specify notions related to relational schema definition and we used the standard \texttt{<AdditionalProperties>} tags to add further notions related to PRMs (e.g., aggregators associated with dependencies, classes associated with nodes).

\subsection{Implemented policies}
\label{policies}
\textbf{Policy for generating the relational schema DAG structure.} To randomly generate the relational schema DAG structure, we use PMMixed algorithm (cf. Section~\ref{BN}), which generates uniformly distributed DAGs in the DAGs space. The structure generated by this algorithm may be a disconnected graph whereas we are in need of a DAG structure containing a single connected component. To preserve this condition together with the interest of generating uniformly distributed examples, we follow the \textit{rejection sampling technique}. The idea is to generate a DAG following PMMixed principle, if this DAG contains just one connected component, then it is accepted, otherwise it is rejected. We repeat these steps until generating a DAG structure satisfying our condition.

\textbf{Policies for generating attributes and their cardinalities.} Having the graphical structure, we continue by generating, for each relation $R$, a primary key attribute, a set of attributes $\mathcal A$, where $card(\mathcal A)-1 \sim Poisson(\lambda=1)$, to avoid empty sets, and for each attribute $A \in \mathcal A$, we specify a set of possible states $\mathcal V (A)$, where $card(\mathcal V(A))-2 \sim Poisson(\lambda=1)$.

\textbf{Policies for generating the dependency structure.} We follow the PMMixed principle to construct a DAG structure inside each class. Then, in order to add inter-class dependencies, we use a modified version of the PMMixed algorithm where we constrain the choice of adding dependencies among only variables that do not belong to the same class.


\subsection{Complexity of the generation process}
We have reported this work to this stage as it is closely related to the choice of the implementation policies. Let $N$ be the number of relations (classes), we report the average complexity of each step of the generation process. 

\textbf{Complexity of the relational schema generation process.} Algorithm~\ref{algogrs} is structured of three loops. Namely, the most expensive one is the first loop dedicated for the DAG structure construction and uses the PMMixed algorithm. Time complexity of the PMMixed algorithm is $\mathcal O(N* \lg N)$. This algorithm is called until reaching the stop condition (i.e., a connected DAG). Let $T$ be the average number of calls of the PMMixed algorithm. $T$ is the ratio of the number of all connected DAG constructed from $N$ nodes~\cite{Robinson1} to the number of all DAGs constructed from $N$ nodes~\cite{Robinson2}. Time complexity of Algorithm~\ref{algogrs} is is $\mathcal O(T*N* \lg N)$.

\textbf{Complexity of the dependency structure generation process.} As for Algorithm~\ref{algogrs}, the most expensive operation of Algorithm~\ref{depStructAlgo} is the generation of the DAG structure inside each class $X_{i\in\{1\ldots N\}}\in \mathcal X$. Through Algorithm~\ref{algogrs}, a set of attributes $\mathcal A(X_{i})$ has been generated for each $X_{i}$. As $card(\mathcal A(X_{i}))-1 \sim Poisson(\lambda=1)$, following Section~\ref{policies}, Then the average number of generated attributes for each class is $lambda=1 +1=2$. Then time complexity of the algorithm is $\mathcal O(N*2*\lg 2)$.

\textbf{Complexity of the slot chains determination process.} The most expensive operation of Algorithm~\ref{SCAlgo} is the $Generate\_Potential\_Slot\_chains$ method. This latter explores recursively the relational schema graph in order to find all paths (i.e., slot chains) of length $k \in \{0\ldots K_{max}\}$. Time complexity of this method is $\mathcal O(N^{K_{max}})$.

\textbf{Complexity of the relational skeleton generation process.} 
The relational skeleton generation algorithm is basically an iteration of depth first search over a relational schema. Thus, complexity of the algorithm would be the same as that of a DFS, i.e. $\mathcal{O}(V+E)$ where $V$ and $E$ are respectively the number of vertices and the number of edges in the graph.


\section{Conclusion and perspectives}
We have developed a process that allows to randomly generate probabilistic relational models and instantiate them to populate a relational database. The generated relational data is sampled from not only the functional dependencies of the relational schema but also from the probabilistic dependencies present in the PRM. 



Our process can more generally be used by other data mining methods as a probabilistic generative model allowing to randomly generated relational data. Moreover, it can be enriched by test query components to help database designers to evaluate the effectiveness of their RDBMS components.

\bibliographystyle{plain}
\bibliography{sample}
\end{document}